%% file: arxiv.tex
\documentclass[10pt,twocolumn,letterpaper]{article}

\usepackage[pagenumbers]{wacv} 
\usepackage[accsupp]{axessibility}  

\input{preamble}

\definecolor{wacvblue}{rgb}{0.21,0.49,0.74}
\usepackage[pagebackref,breaklinks,colorlinks,allcolors=wacvblue]{hyperref}
\usepackage{pifont}
\usepackage{multirow}
\def\confName{WACV}
\def\confYear{2026}

\title{MMHOI: Modeling Complex \D{3D Multi-Human} Multi-Object Interactions}

\author{Kaen Kogashi\\
Mitsubishi Electric\\
Japan
\and
Anoop Cherian\\
Mitsubishi Electric Research Labs\\
United States
\and
Meng-Yu Jennifer Kuo\\
Nara Women's University\\
Japan\\
}

\begin{document}
\twocolumn[{
\maketitle
\centering
\vspace{-20pt}

\begin{center}
      \centering
    \includegraphics[clip, trim=0.0cm 0.0cm 0.0cm 0.0cm, width=1\linewidth]
    {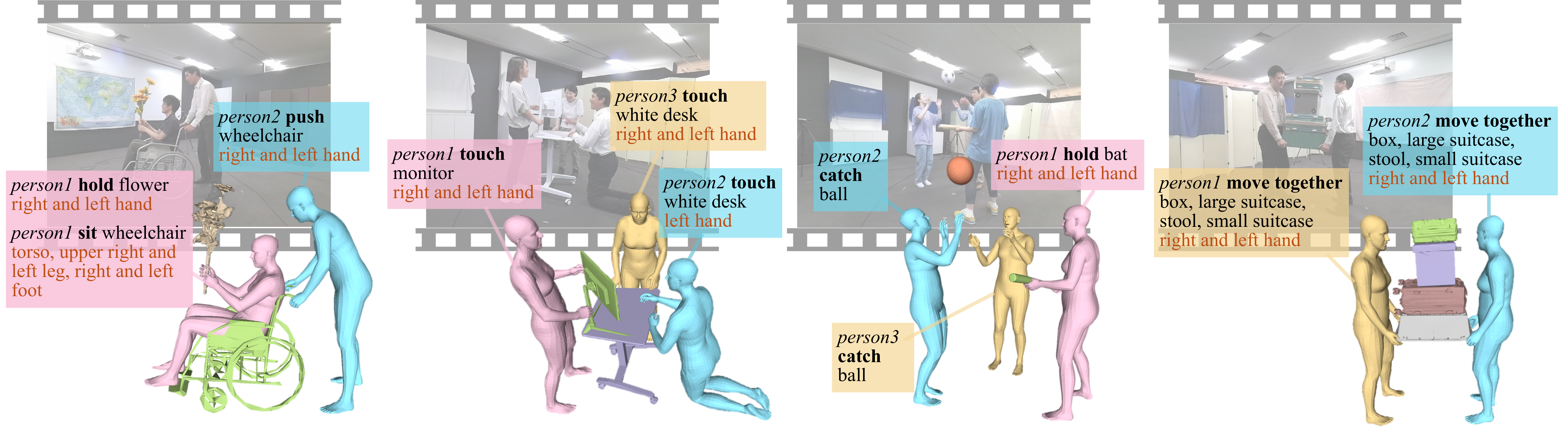}
    \vspace{-23pt}
    \captionof{figure}{Example scenes from our MMHOI dataset -- a new, large-scale dataset with high-quality annotations of multiple 3D humans, objects, actions, and interaction body parts, enabling holistic reasoning in complex interaction scenarios. Person IDs are shown in \textit{italic}, action annotations in \textbf{bold}, object names in regular font, and interacting body parts are highlighted in \textcolor{RedOrange}{orange text}.
    }
\label{fig:Intro}
\end{center}	
}]

\begin{abstract}
\vspace*{-18pt}

Real‑world scenes often feature multiple humans interacting with multiple objects in ways that are causal, goal‑oriented, or cooperative. Yet existing 3D human-object interaction (HOI) benchmarks consider only a fraction of these complex interactions. To close this gap, we present MMHOI -- a large-scale, Multi-human Multi-object Interaction dataset consisting of images from 12 everyday scenarios. MMHOI offers complete 3D shape and pose annotations for every person and object, along with labels for 78 action categories and 14 interaction‑specific body parts, providing a comprehensive testbed for next-generation HOI research. 

Building on MMHOI, we present MMHOI-Net, an end-to-end transformer-based neural network for jointly estimating human–object 3D geometries, their interactions, and associated actions. A key innovation in our framework is a structured dual-patch representation for modeling objects and their interactions, combined with action recognition to enhance the interaction prediction. Experiments on MMHOI and the recently proposed CORE4D datasets demonstrate that our approach achieves state-of-the-art performance in multi-HOI modeling, excelling in both accuracy and reconstruction quality. The MMHOI dataset is available at \projectlink{https://zenodo.org/records/17711786}.
\end{abstract}

\section{Introduction}
Perceiving 3D human-object interactions (HOIs) is central to understanding human behavior and is crucial for applications like action recognition, 3D scene reconstruction, and human-centric AI.
While progress has been made in modeling 3D interactions between a single human and a single object \cite{bhatnagar2022behave, fan2023arctic, huang2024intercap, GRAB:2020, jiang2023chairs, mandery2016unifying, zhang2023neuraldome, cseke_tripathi_2025_pico}, real-world scenarios often involve multiple humans interacting with multiple objects simultaneously. 
\D{These settings introduce greater complexity due to increased interaction patterns, diverse object categories, and ambiguity in recognizing multi-entity actions.
Addressing these challenges is essential for developing robust models that generalize to the dynamic, collaborative nature of real-world environments.} 

\D{In response, recent efforts have begun exploring more complex interaction settings beyond single-human scenarios. One such effort is HOI-M$^3$ \cite{zhang2024hoi}, which expands the scope to multiple humans and objects. However, it treats humans and objects as independent entities, overlooking the collaborative dynamics of real-world interactions. Recently, CORE4D \cite{liu2025core4d} complements this by capturing 3D interactions with cooperative intent, but focuses only on two-person-single-object scenarios with limited action diversity.
Moreover, the lack of varied action and detailed body part annotations limits these datasets' ability to provide high-level task-specific guidance for low-level interaction modeling, leaving the synergy between 3D reconstruction and action recognition in multi-HOI scenarios largely unexplored.} 

To address the above challenges, we introduce \textbf{MMHOI}, a novel large-scale dataset that provides comprehensive 3D annotations for multiple humans and objects, coupled with action labels that capture diverse real-world interactions. See Fig.~\ref{fig:Intro},~\ref{fig:overview_MMHOI} for a few example scenes from our dataset. MMHOI offers a combination of spatial and semantic data, consisting of \D{$\sim$600k frames}
captured across 12 daily-living scenarios, featuring $13$ participants ($7$ males and $6$ females) interacting with $22$ commonly used objects. Each interaction is annotated with 78 distinct action classes, enabling the study of both cooperative and non-cooperative activities. Unlike previous datasets, MMHOI incorporates $14$ interactive body part labels that reveal the specific body regions involved in object interactions, facilitating a deeper understanding of human-object interaction dynamics at both the spatial and semantic levels.

Building upon the MMHOI dataset, we propose a novel single-shot framework that simultaneously estimates the pose, shape, and 3D location of all humans and objects in camera space from a single RGB image. Human meshes are parameterized using the SMPL-X model \cite{SMPL-X:2019}, while object poses are represented with 6DoF parameters. Our framework leverages a Vision Transformer (ViT)-based architecture \cite{dosovitskiy2020vit, multi-hmr2024} to extract patch-level features for both human and object modeling. To capture object-specific spatial and interaction cues, we introduce a structured dual-patch representation: each object is described by a main patch covering its primary region and a sub-patch encoding interaction-relevant features. This representation enables the model to reason about both object properties and fine-grained human-object interactions more effectively.

Another key aspect of our framework is the use of action recognition as a signal to enhance 3D reconstruction. Specifically, we use the features from the human and object
perception modules in our transformer network to predict action labels and the  interaction body parts. Next, body-part interaction consistency loss constraints are enforced between interactive body parts' ground truth classes and corresponding object regions, improving spatial alignment and preventing physically implausible configurations.

We present experiments demonstrating the benefits of our method in multi-human multi-object interactions modeling on our proposed MMHOI and the recent CORE4D datasets. We benchmark our model against prior approaches that are re-purposed to our task setting. Our results show that while standard methods falter in inferring cross-interactions, our method excels in modeling complex interactions, achieving state-of-the-art results and superior reconstruction of humans, objects, and their interactions. Our key contributions are summarized below.

\begin{itemize}
    \item \D{We introduce \textbf{MMHOI}, a novel large-scale dataset featuring detailed multi-HOI 3D annotations, action labels, and interactive body part labels. To the best of our knowledge, MMHOI is the only dataset comprising simultaneous multi-human and object-object  interactions.}

    \item \D{We propose MMHOI-Net, a novel method for multi-human and multi-object interaction reconstruction in 3D, incorporating a structured dual-patch object representation and leveraging action guidance.} 

    \item \D{The effectiveness of MMHOI-Net is demonstrated through extensive experiments on MMHOI, and also on the recent CORE4D \cite{liu2025core4d} datasets.}
\end{itemize}


\section{Related Works}
\noindent\textbf{3D Single Human and Object Interaction.} Several works that focusing on full-body interactions have been proposed to advance the HOI modeling \cite{bhatnagar2022behave, jiang2023chairs, mandery2016unifying, GRAB:2020, zhang2023neuraldome, cseke_tripathi_2025_pico}. For instance, the BEHAVE dataset \cite{bhatnagar2022behave} benchmarks full-body human interactions with a single movable object, facilitating progress in modeling isolated HOI tasks \cite{xie2022chore, nam2024joint}. 
However, these datasets predominantly focus on interactions involving a single human and a single object, overlooking the more complex scenarios where multi-human interact with multi-object in cooperative environments.

\noindent\textbf{3D Multiple Humans and Objects Interaction.} 
Recently, the HOI-M$^3$ dataset \cite{zhang2024hoi} advanced the field by capturing interactions among multi-humans and multi-objects in shared contextual environments. 
However, it lacks scenarios in which multiple humans simultaneously interact with one or more objects, leaving a significant gap for the explicit studying of cooperative interactions. 
\D{More recently, CORE4D \cite{liu2025core4d} focuses on two-person interactions involving a single object, capturing aspects of cooperation however has limited action annotations and do not consider multi-object scenarios. 
The absence of detailed action and interactive body part annotations in both datasets limits their scope in modeling interaction consistency and physical plausibility.
Instead, our MMHOI dataset captures rich, complex interactions involving co-existing multi-human and multi-object-object interactions, with diverse action labels.}

\noindent\textbf{Action Recognition and 3D HOI.} Action recognition and 3D human pose estimation are deeply interconnected. Early works primarily relied on 2D keypoints for action recognition \cite{TIN-net,PMF}, however recent research has demonstrated that 3D pose estimation provides rich geometric information and context for complex interaction reasoning~\cite{luvizon2020multi}. Although sparse 3D body joint representations often fail to capture fine-grained details, detailed 3D poses \cite{rajasegaran2023benefits,li2020detailed} are seen to improve action recognition and single HOI modeling. Leveraging these insights, we propose to use action recognition to help enforce consistency in multi-HOI 3D estimation, ensuring that the reconstructed body poses remain faithful to their interaction objects.


\section{MMHOI Dataset}

The MMHOI dataset is designed to capture multi-HOIs among cooperative interactions, providing high-quality data for advancing 3D reconstruction and action understanding. In \tabref{table:comparison}, we show a comparison between MMHOI and prior datasets. Below we outline the key statistics, data acquisition, and processing pipeline \textbf{(see supp. for details)}.

\subsection{Dataset Statistics}
\begin{figure}[t]
\begin{center}
\includegraphics[clip, trim=0.0cm 0.0cm 0.0cm 0.0cm, width=1\linewidth]
{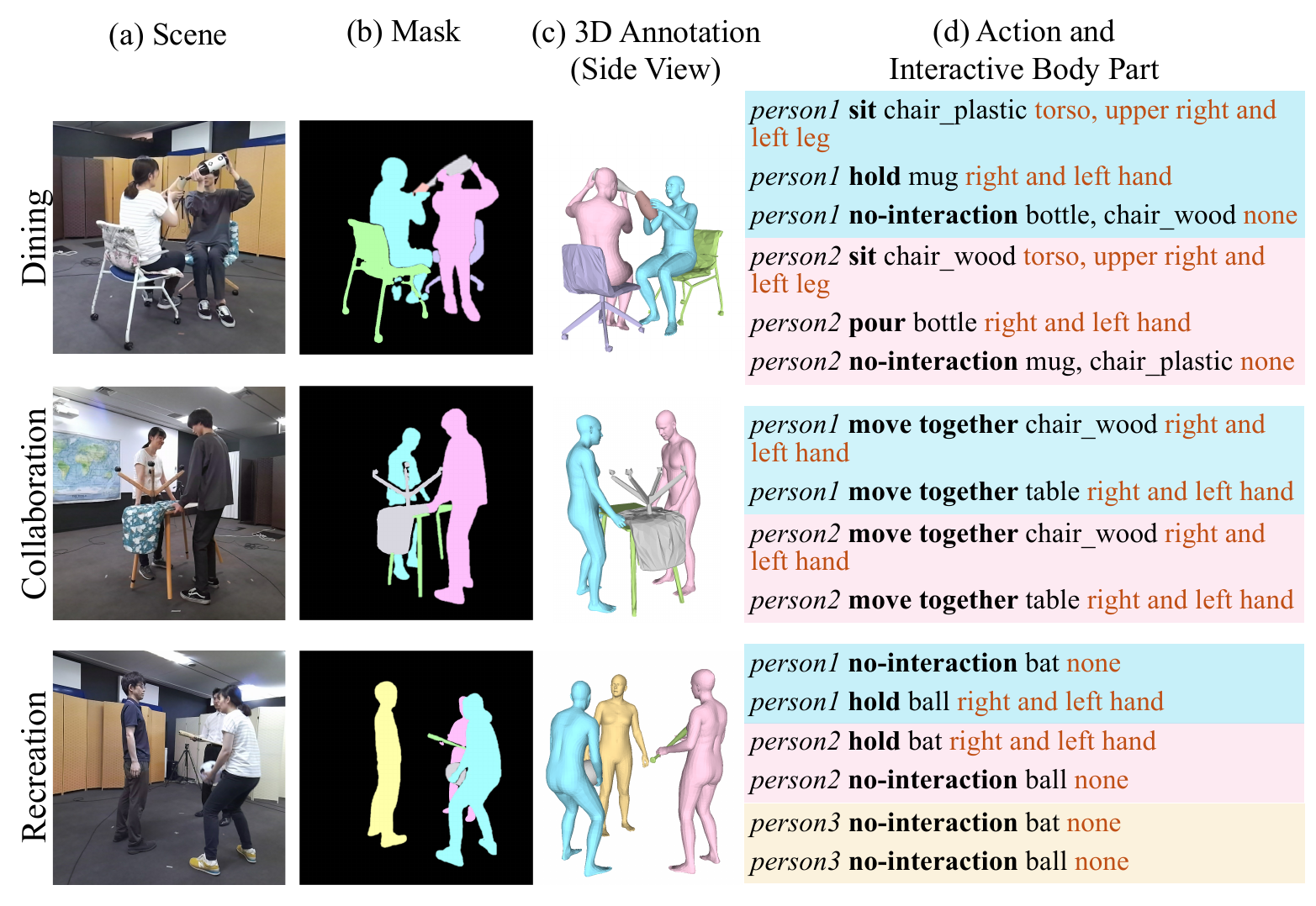}
\end{center}
\vspace{-13pt}
   \caption{Overview of MMHOI. The dataset is categorized into three main interaction types: dining, collaborative work, and recreational activities, each type belongs to 12 scenarios. MMHOI consists of (a) RGB images, (b) segmentation masks, (c) 3D tracking of multiple humans and objects, and (d) action and interactive body part labels.} 
\label{fig:overview_MMHOI}
\end{figure}

Our dataset captures realistic and diverse human-object interactions in 12 daily scenes, such as dining, collaborative work, and recreational activities (see \figref{fig:overview_MMHOI}). It comprises
\D{$\sim$600k frames} featuring interactions performed by 13 participants (7 males and 6 females) with 22 commonly used objects spanning various categories.
To complement the 3D and object annotations, MMHOI includes action annotations covering 78 distinct action classes and $14$ interacting body parts.
Notably, MMHOI is the first dataset to combine 3D annotations of multi-human and multi-object-object with detailed action annotations, capturing their interactions in diverse and realistic scenarios.\footnote{Each subject in our dataset provided their written consent to be included.}

\subsection{Dataset Acquisition}
The MMHOI data was captured using a multi-camera setup consisting of four Microsoft Azure Kinect sensors. Each sensor recorded scenes at a resolution of 2048 $\times$ 1536 pixels and 30 fps. Calibration across the four sensors was performed using AprilTag markers \cite{song7extrinsic} to ensure precise spatial alignment. Although the Azure Kinect cameras provide built-in synchronization, occasional data drift due to hardware limitations necessitated manual adjustments to achieve accurate temporal alignment.

\subsection{Data Processing}
To achieve high-quality annotations, a rigorous pipeline was developed to process humans, objects, and actions within the scenes.
\vspace{-10pt}
\paragraph{Human Annotations.}  
Using synchronized and calibrated four-view videos, human segmentation was performed with the L-SAM \cite{lsam}. 
We employ SMPL-X \cite{SMPL-X:2019} as the 3D parametric human model to represent the statistical variations in human shape and pose. Following \cite{multi-hmr2024}, the model parameters include pose parameters $\theta \in \mathbb{R}^{53 \times 3}$ and shape parameters $\beta \in \mathbb{R}^{10}$. From these parameters, SMPL instantiates a human body mesh $M \in R^{N\times3}$ with $N = 10, 475$ vertices.
To initialize 3D human body representations from each camera view, we employed Multi-HMR \cite{multi-hmr2024}, a single-view model that estimates SMPL-X parameters independently for each viewpoint.
Professional annotators reviewed the results, selecting the view with minimal occlusion for refinement. These models were further refined using Iterative Closest Point (ICP) alignment with depth data (recorded from Kinect) in 3D space, followed by manual adjustments with FreeCAD \cite{riegel2016freecad} to finalize their positions. 

\vspace{-10pt}
\paragraph{Object Annotations.}  
\D{Object segmentation was also performed using L-SAM \cite{lsam}. All objects in MMHOI are pre-scanned with the FreeScan UE Pro2 to obtain high-fidelity CAD models. These are initially aligned to depth maps (recorded from Kinect) via ICP to estimate 6D poses, then manually refined by annotators for accurate placement. We set the center of four Kinects as the world origin, and the CAD model's center as the object’s origin. The object’s final pose ($R$, $T$) is computed via ICP between the original CAD template and the manually annotated placement.}

\vspace{-10pt}
\paragraph{Action Annotations.}  
\D{In addition to 3D annotations of humans and objects, MMHOI provides detailed action annotations to contextualize interactions. Each frame is labeled by trained annotators, who assign predefined action classes and specify which of the 14 body parts are involved in the interaction. The annotations are inspired by daily scenarios in video-based action datasets \cite{AVA, CMU_Panoptic, Hollywood2, Moments, TV_Human_Interaction, SALSA, UT-Interaction, Kinetics700, shakefive, SBU_Kinect, HACS}, and reflect the concept of \emph{synergy}—how human actions and object affordances complement each other. Actions include both cooperative and individual tasks, such as “passing an object,” “assembling,” and “organizing.” Annotators cross-referenced multi-view RGB-D recordings to ensure accurate alignment between actions and 3D spatial configurations. Action labels are formatted as time-aligned sequences, enabling seamless integration with 3D and semantic data.}

\begin{table}[t]
\setlength\tabcolsep{2pt}
\caption{Comparisons between MMHOI and 3D HOI reconstruction datasets. MMHOI includes inter- and intra- human-and-object interactions, along with action and interaction body part annotations. 
 We include the maximum \# of humans per scene, the \# of (conc)urrent interactions, and the \# of interacting parts. * indicates captions are provided instead of action labels.
}
\vspace{-15pt}
\label{table:comparison}
\centering
\begin{center}
\resizebox{8.5cm}{!}{
\begin{tabular}{|l|c|c|c|c|c|c|c|}
\hline
\small {Datasets} &   \small{ \makecell{Dyn.\\Obj.}}& \small{\makecell{\D{\# Conc.}\\Interact.}} & \small{\makecell{\# Interact. \\Parts} }& \small{\makecell{\D{Interact}.\\Action}} & \small{\makecell{max. \#\\Humans} }& \small{\makecell{\D{Obj-Obj}\\Interact.}}
\\
\hline\hline
GRAB \cite{GRAB:2020}
&  \ding{51}& \ding{55} &\ding{55} &\ding{55} &\ding{55} &\ding{55}\\
BEHAVE \cite{bhatnagar2022behave}& \ding{51}& \ding{55} &\ding{55} &\ding{55} &\ding{55} &\ding{55}\\
CHAIRS \cite{jiang2023chairs}& \ding{51}&\ding{55}&\ding{55}&\ding{55} &\ding{55} &\ding{55}\\
NeuralDome \cite{zhang2023neuraldome}& \ding{51}& \ding{55} &\ding{55} &\ding{55} &\ding{55} &\ding{55}\\
InterCap \cite{huang2024intercap}&\ding{51}&\ding{55} &\ding{55}&\ding{55} &\ding{55}&\ding{55} \\
HIMO \cite{lv2024himonewbenchmarkfullbody}&  \ding{51}&\ding{55}&\ding{55}&\ding{55} &\ding{55} & \ding{51}\\
\D{ParaHome} \cite{kim2024parahome} &  \ding{51}&\ding{55}&\ding{55}&\ding{51}*&\ding{55} &\ding{51}\\
HOI-$\mathrm{M}^3$ \cite{zhang2024hoi}& \ding{51} &\ding{55}&\ding{55}&\ding{51}* &\D{$\geq$3}&\ding{55}\\
CORE4D \cite{liu2025core4d}& \ding{51} &1&\ding{55}&  5 &2& \ding{55}\\
\textbf{MMHOI (Ours)}& \ding{51} &4&14&  21&3&\ding{51}\\
\hline
\end{tabular}
}
\end{center}
\end{table}

\vspace{-10pt}
\section{MMHOI-Net}
This section introduces our MMHOI-Net, a vision-transformer based single-shot multi-task model for jointly learning the 3D human-object configurations, inferring the actions, and refining the body-part interactions. In Fig.~\ref{fig:overview_method}, we illustrates the overall architecture of MMHOI-Net. The input to our model is a single RGB image $\mathbf{I} \in \mathbb{R}^{H \times W \times 3}$ of resolution $H \times W$ depicting a multi-HOIs scene and a mask of this scene produced using L-SAM~\cite{lsam}. Our model outputs the SMPL-X parameters of all individual humans in $\mathbf{I}$, all the object meshes in the scene,   their corresponding root 3D locations $t \in \mathbb{R}^3$ in camera coordinates, and associated action classes between all humans and object pairs.

\subsection{Vision Transformer-Based Feature Encoding}

\begin{figure*}[t]
\begin{center}
\includegraphics[clip, trim=0cm 0cm 0cm 0cm, width=1\linewidth]{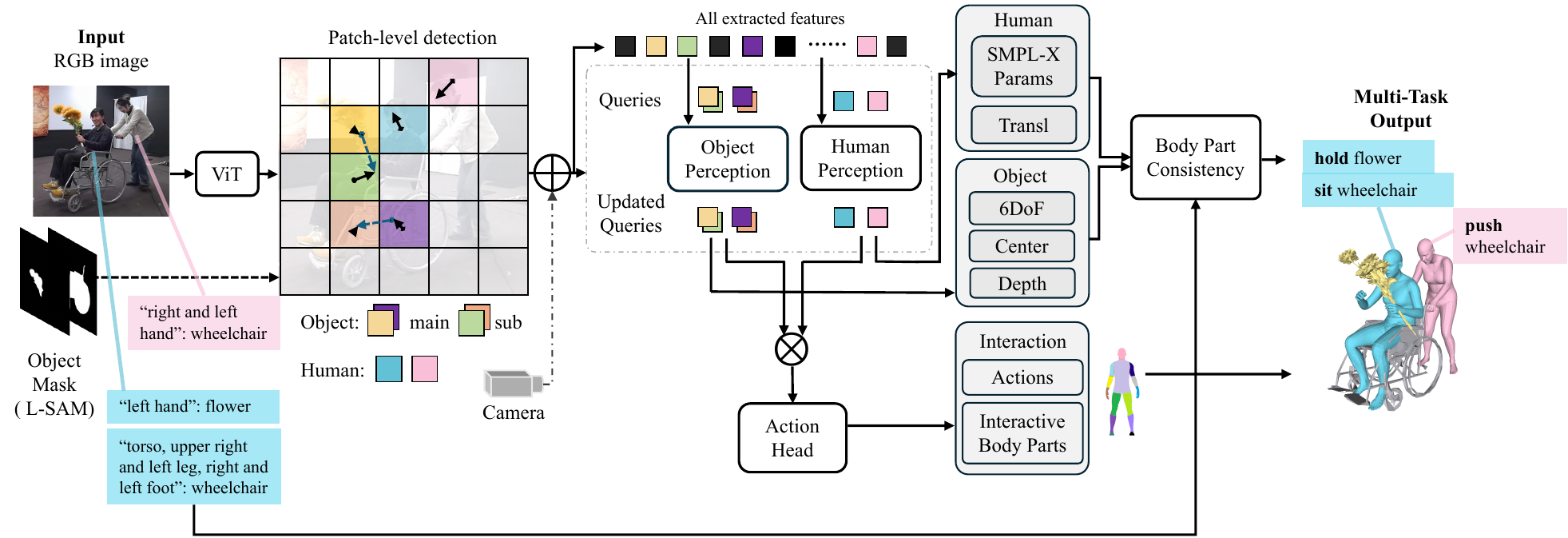}
\end{center}
\vspace{-15pt}
\caption{MMHOI-Net model architecture. Given a single RGB image, our model jointly estimates the 3D geometry of multiple humans and objects while incorporating action recognition as a supervisory signal. A ViT backbone extracts patch-level features. For human perception, detected keypoints serve as queries in a Human Perception Head \cite{multi-hmr2024}, regressing SMPL-X pose, shape, and translation. Object perception head uses a structured dual-patch representation to regress \D{object 6DoF, center, and depth.} An action MLP predicts action and interaction body part classes.}
\label{fig:overview_method}
\end{figure*}

To effectively process the input image, we employ a Vision Transformer (ViT) backbone \cite{dosovitskiy2020vit} to extract high-level image features. 
The image $\mathbf{I}$ is first subdivided into non-overlapping patches of size $P \times P$. 
Each patch is embedded into a feature token, and these tokens are processed with self-attention blocks into $\mathbf{E} \in \mathbb{R}^{H/P \times W/P \times D}$ with dimension $D$ features. 
Similar to Multi-HMR~\cite{multi-hmr2024}, each output token maps spatially to a patch of the input RGB image. We employ the Human Perception Head (HPH)~\cite{multi-hmr2024} for modeling the humans in the scene, and propose an Object Perception Head (OPH) for modeling the mesh parameters of the objects. Both these heads use the same ViT backbone, which is initialized with weights from a pre-trained MultiHMR model~\cite{multi-hmr2024} during training.  Below, we provide more details of HPH and our novel OPH. As depicted in Fig.~\ref{fig:overview_method}, we further concatenate features from HPH and OPH for the action and body part classification.

\subsection{Human Perception Head}
Following Multi-HMR~\cite{multi-hmr2024}, this head identifies all major human keypoints and predicts their SMPL-X parameters and depth. For $N$ humans in an image, a set of queries $\{\mathbf{q}_n\}_{n\in[N]}$ is initialized from the ViT feature tensor and processed in parallel through a stack of $L$ blocks ($L=2$ in the experiments), where alternating cross-attention and self-attention layers to progressively refine the queries. The final output, $\mathbf{Q}^L \in \mathbb{R}^{(D+D') \times N}$, is a set of $N$ refined human features used to regress the final human parameters (see  \cite{multi-hmr2024} for further details).

\subsection{Object Perception Head}
\D{To jointly regress human and object geometries, we introduce an \textit{Object Perception Head (OPH)} in our model (\figref{fig:overview_method}). Similar to the HPH branch for humans, OPH for objects follow a similar structure, however incorporates two patches instead of one -- a main patch and a sub-patch -- to model objects at multiple scales, which we call a \emph{structured dual-patch representation}. Why do we need two patches?  Recall that in~\cite{multi-hmr2024}, the human head is used as the primary keypoint to place the reconstructed human mesh. Objects in our dataset, however, vary in shape and symmetry, and scale. This is further exacerbated across categories, and thus treating them as monolithic entities is inadequate. 
To overcome this, we propose to use two patches overlapping the object mask to more robustly capture the object’s location, pose, and interaction points.}

\begin{figure}[t]
\begin{center}
\includegraphics[clip, trim=0cm 0cm 0cm 0cm, width=1\linewidth]{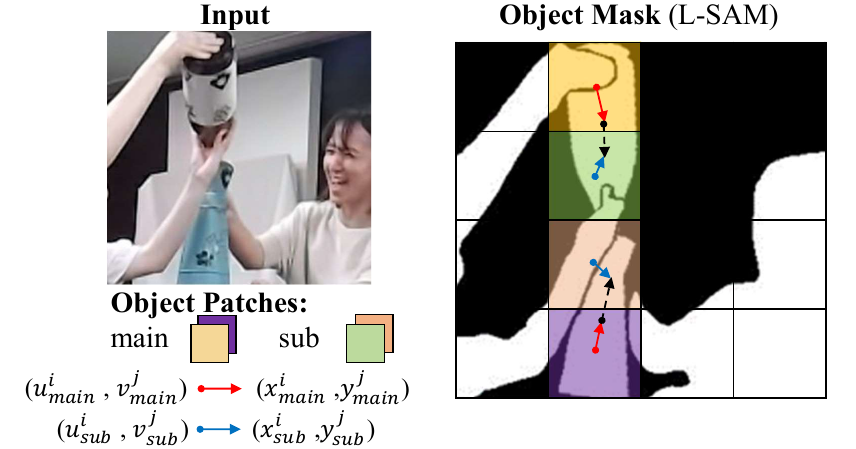}
\vspace*{-20pt}
\end{center}
   \caption{Our structured dual-patch object representation for inferring the object mesh parameters. The black arrows indicate approximate orientations of the objects. 
   }
\label{fig:object_patches}
\vspace*{-15pt}
\end{figure}

Our dual patch representation is obtained as follows. Given an object mask detected using L-SAM \cite{lsam}, the main patch for the object is defined as a patch covering the largest region of the mask containing the object mask center. The sub-patch is selected from one of the 8 surrounding patches to the main patch, corresponding to the second-largest region of the object mask and containing human-object interaction masks, i.e., patches containing mask parts belonging to both humans and the object (see \figref{fig:object_patches}).\footnote{If no such interaction patches are found, we use the second largest patch containing the object mask.}

In order for our model to be aware of the pose of the object, we propose to regress local patch offsets that implicitly guide the MMHOI-Net towards inferring the holistic object pose using the dual-patch representation. Let the main patch coordinates, \ie, the object center $(x_{main},y_{main})$, be defined as the center of the \D{smallest} bounding box encapsulating the L-SAM produced object mask.  Suppose for a main patch at location $(i,j)$, if $(u_{main}^i, v_{main}^j)$ is the patch center, then we define the offsets $ (\delta_{main_u}^i, \delta_{main_v}^j) := (x^i_{main} -u_{main}^i, y^j_{main} - v_{main}^j)$. As noted by the red direction vectors in the main patches (yellow and purple in Fig.~\ref{fig:object_patches}), this offset is pointed from the patch center to the object mask centroid within the main patch. The sub-patch center coordinates $(x_{sub}, y_{sub})$ are computed in a similar manner to the main patch (except that this point is the centroid of the object mask within the sub-patch)  using the second largest mask region per patch (pink and green patches in Fig.~\ref{fig:object_patches}). For the sub-patch center $(u_{sub}^i, v_{sub}^j)$, the sub-patch offset coordinates $(\delta_{sub_u}^i, \delta_{sub_v}^j )$ are  computed as $(\delta_{sub_u}^i, \delta_{sub_v}^j) :=  (x_{sub}^i - u^i_{sub}, y_{sub}^j - v^j_{sub})$. As is clear, the sub-patch offsets point from the sub-patch center to the sub-patch mask center (depicted by the blue arrows in Fig.~\ref{fig:object_patches}). The main and the sub-patch offsets jointly capture a ray pointed from the object center in the direction of the sub-patch, characterizing an approximate orientation of the object (black arrows). Thus, our key insight is that by having the network to regress these offsets during training (through a suitable loss that will be discussed shortly), it implicitly learns the pose of the object mesh.

Similar to the processing pipeline in the HPH head, given $M$ objects in an image (detected by L-SAM), the final output of the transformer cross-attention layers is given by $\mathbf{Q}^{2 \times L} \in \mathbb{R}^{2 \times (D+D' \times M)}$ and viewed as a set of $2 \times M$ output object features. Next, the updated main and sub-patch queries from OPH are concatenated, and regressed using an MLP to produce the 6D object pose, center, and depth in 3D. To further enhance localization accuracy, available camera parameters are embedded into each patch as in \cite{multi-hmr2024}.

\subsection{Human-Object Interaction (Action) Head}
\label{sec:hoi_head}
\D{We observe that action cues offer valuable semantic context, revealing the underlying structure of 3D human-object interactions. For instance, if an action module predicts ``picking up a box'', the corresponding body parts (\eg, hands) must be touching the box. Inspired by prior works~\cite{nam2024joint, xie2022chore} that use interactive body parts for 3D reconstruction in single-HOI, we extend this to multi-HOI by introducing an HOI head for action detection (see \figref{fig:overview_method}).}
To this end, we do a pairwise concatenation of the set of $N$ human features output from HPH and a set of $2 \times M$  object features output from OPH, thus producing features $G=\{H_{main}\}_N \otimes \{(O_{main} \oplus O_{sub})\}_M $, where $\{H_{main}\}_N$ denotes the set of $N$ main patch (head) output features from the HPH branch and $\{(O_{main} \oplus O_{sub})\}_M$ indicate the concatenation of the output object main $O_{main}$ and sub-patch $O_{sub}$ features  from OPH, respectively.  After this step, we obtain $|G|=N\times M$ human-object pairs from the direct product, which is passed to a 2-layer MLP to jointly predict both action and interactive body part classes.

\subsection{Training Losses}
Our model is optimized through a weighted combination of human and object reconstruction, interaction prediction, body-part detection, and 3D HOI reconstruction losses. 

\noindent\textbf{Human Reconstruction Loss.} 
Similar to~\cite{multi-hmr2024}, the updated queries of the HPH branch are viewed as a set of $N$ output features, which are used to regress the $N$ human parameters using a shared MLP.
The reconstruction loss aims to ensure accurate recovery of 3D human geometries in still images. The overall loss is defined as:
\begin{align}
    \mathcal{L}_{hum}\!=\!\lambda_{h}(\mathcal{L}_{hproj}\!+\!\mathcal{L}_{hmesh})+\!\! 
    \lambda_{param}\mathcal{L}_{param}\!+\!\lambda_{det}\mathcal{L}_{det},\notag
\label{eq: reconst_loss}
\end{align}
where $\mathcal{L}_{hproj}$ and $\mathcal{L}_{hmesh}$ minimizes the re-projection error and output meshes for humans. $\mathcal{L}_{param}$ regularizes the SMPL pose, shape parameters, human depth, and human patch offset regression. The $\mathcal{L}_{det}$ captures the human detection loss.  $\mathcal{L}_{hproj}$, $\mathcal{L}_{hmesh}$, and $\mathcal{L}_{param}$ minimize with $L_{1}$ regression losses. $\mathcal{L}_{det}$ is minimized using binary cross-entropy loss. $\lambda_{h},\lambda_{param}, \lambda_{hdet}$ are the loss weights. Refer to supp. materials for details.

\vspace{-10pt}
\paragraph{Object Reconstruction Loss.} 
The final output of the OPH branch are viewed as a set of $2\times M$ output features and is used to regress the $M$ object parameters using a shared MLP. 
The object reconstruction loss aims to ensure accurate recovery of the 3D object geometries in still images. The overall loss is defined as:
\begin{equation}
\mathcal{L}_{obj}\!=\!\lambda_{o}(\mathcal{L}_{oproj}+\mathcal{L}_{omesh})+\lambda_{p}\mathcal{L}_{p}+\lambda_{main} \mathcal{L}_{main}+\lambda_{sub} \mathcal{L}_{sub},\notag
\label{eq: off_loss}
\end{equation}

\noindent where the object related loss $\mathcal{L}_{p}$ refers to inferring the 6D object pose, i.e., predicting the translation $T \in \mathbb{R}^{3}$, rotation $R \in \mathbb{R}^{3\times3}$, center $C \in \mathbb{R}^{3}$ (as in~\cite{xie2022chore}) and depth (m) in 3D space. The input mesh to $\mathcal{L}_{omesh}$ input is computed by transforming the object mesh using the predicted $R$ and $T$ from the original CAD coordinates. $\mathcal{L}_{oproj}$ computes the 2D re-projection error of the 3D object mesh. $\mathcal{L}_{p}$ minimizes errors in object rotation, translation, center, and depth \wrt the ground truth. The main and sub-patches offset regression losses captured in $\mathcal{L}_{main}$ and $\mathcal{L}_{sub}$ minimize the predicted main and sub-patch offset coordinates $(\delta_{main_u}^i, \delta_{main_v}^j )$ and $(\delta_{sub_u}^i, \delta_{sub_v}^j )$ to the GT offset. All of $\mathcal{L}_{oproj}$, $\mathcal{L}_{omesh}$, $\mathcal{L}_{p}$, $\mathcal{L}_{main},$ and $ \mathcal{L}_{sub}$ are minimized with $L_{1}$ losses. $\lambda_{o},\lambda_{p}, \lambda_{main},$ and $\lambda_{sub}$ are the loss weights. 

\vspace{-10pt}
\paragraph{Interaction Loss.}  
The interaction loss enforces semantic and geometric consistency in human-object interactions by refining 3D reconstruction based on action and interactive body part detections. It is defined as:
\begin{equation}
    \mathcal{L}_{interact} =  \lambda_{act}\mathcal{L}_{act} + \lambda_{bp}\mathcal{L}_{bp} + \lambda_{cons}\mathcal{L}_{cons}.\notag
\end{equation}
Inputs to $\mathcal{L}_{act}$ and $\mathcal{L}_{bp}$ losses are the predicted action and body part classes from the \D{HOI (action)} head. Input to the $\mathcal{L}_{cons}$ is the human body part and object 3D meshes from HPH and OPH, respectively, as well as the body part GT labels.
We use the cross-entropy losses in \(\mathcal{L}_{act}\) and \(\mathcal{L}_{bp}\) for action and interactive body part classification, respectively. The consistency loss \(\mathcal{L}_{cons}\) aligns predicted interactive body parts with their corresponding object regions to prevent physically implausible configurations. Specifically, to enforce alignment, \(\mathcal{L}_{cons}\) minimizes the distance between the ground-truth interactive body part points set \(\mathcal{P}^i_\mathrm{bp}\), where \(i \in \mathcal{I}\subseteq \{1, \dots, N\}\) indexes body parts and the predicted object points set \(\mathcal{P}_\mathrm{o}\):
\begin{equation}
    \mathcal{L}_\mathrm{cons} = \sum_{i\in\mathcal{I}} \max\left( \Psi(\mathcal{P}^i_\mathrm{bp}, \mathcal{P}_\mathrm{o})- \delta, ~0\right),\notag
\end{equation}
\noindent where \(\Psi(\mathcal{P}^i_\mathrm{bp}, \mathcal{P}_\mathrm{o})\) is the Chamfer Distance between \(\mathcal{P}^i_\mathrm{bp}\) and \(\mathcal{P}_\mathrm{o}\), with a threshold \(\delta = 5\mathrm{mm}\). If \(\Psi\) exceeds \(\delta\), the loss encourages proximity.  $\lambda_{act}, \lambda_{bp}, \lambda_{cons}$ are the loss weights. Our  final training objective is as follows:
\begin{equation}
    \mathcal{L}=\mathcal{L}_{hum}+ \mathcal{L}_{obj} + \mathcal{L}_{interact}.\notag
\end{equation}

\begin{table*}[t]
\begin{center}
\setlength\tabcolsep{6pt} 
\caption{State-of-the-art comparisons on MMHOI and \D{CORE4D (S1) test sets. Chamfer Distance (CD)} and Vertex-to-Vertex (V2V) distance are reported in cm for both human and object reconstructions. 
By ``Action'', we denote results obtained by integrating the method with actions. By ``S''/``M'', we mean to apply Procrustes alignment at the single/multi HOI level, i.e., aligning the predicted SMPL-X human meshes and object meshes at a single-HOI level or all at once, respectively. 
} 
\label{table:total_accuracy}
\begin{tabular}{|c|c|c|c|c|c|c|c|}
  \hline
   Dataset &Method & \makecell{S Hum \\CD $\downarrow$} & \makecell{S Obj \\CD $\downarrow$} & \makecell{M Hum \\ CD $\downarrow$} & \makecell{M Hum \\V2V $\downarrow$} & \makecell{M Obj \\ CD $\downarrow$} & \makecell{M Obj  \\ V2V $\downarrow$}
  \\
  \hline
     \multirow{7}{*}{MMHOI} & Multi-HMR \cite{multi-hmr2024}+PHOSA\cite{zhang2020phosa} & 7.50& 82.07 &-  &-  &- &-  \\ 
  &Multi-HMR \cite{multi-hmr2024}+CHORE\cite{xie2022chore}& 7.41& 79.46& - & - & - & - \\
  &Multi-HMR \cite{multi-hmr2024}+CONTHO\cite{nam2024joint}& 7.40& 78.06& - &  - & - & - \\
        \cline{2-8}
       &Multi-HMR \cite{multi-hmr2024}+PHOSA\cite{zhang2020phosa} + Action &7.25 &73.48& 9.41 &  28.81&27.97 & 56.64 \\
  &Multi-HMR \cite{multi-hmr2024}+CHORE\cite{xie2022chore}+ Action & 7.15& 69.94&9.35 &  28.73&26.87 & 53.74 \\
  &Multi-HMR \cite{multi-hmr2024}+CONTHO\cite{nam2024joint}+ Action &7.03 &65.42 &9.33 &  28.71&26.64 & 53.63 \\

   &Multi-HMR \cite{multi-hmr2024}&- & -&9.30 & 28.74  &- &-  \\
    &\textbf{MMHOI-Net (ours)} &\textbf{6.57}& \textbf{55.06}& \textbf{6.47} &\textbf{23.18 }  &  \textbf{21.02} & \textbf{49.85}   \\
      \hline \hline

             &Multi-HMR \cite{multi-hmr2024}+PHOSA\cite{zhang2020phosa} + Action &21.87 & 173.64& 29.74 &83.69  & 81.70 &140.59  \\
  &Multi-HMR \cite{multi-hmr2024}+CHORE\cite{xie2022chore}+ Action & 20.32&161.97 & 28.74&81.85  & 79.32&  134.62\\
  
        CORE4D&Multi-HMR \cite{multi-hmr2024}+CONTHO\cite{nam2024joint}+ Action &19.71 &153.18 &27.15 &  79.78&77.26 & 126.63 \\
   (S1)&Multi-HMR \cite{multi-hmr2024}&- & -&26.85 & 76.74  &- &-  \\
    &\textbf{MMHOI-Net (ours)} &\textbf{18.40}& \textbf{144.17}& \textbf{18.11} &\textbf{57.95}  &  \textbf{58.82} & \textbf{119.61}   \\ \hline
\end{tabular}
\end{center}
\vspace*{-20pt}
\end{table*}

\section{Experiments and Results}
\paragraph{Evaluation Datasets.} \AC{We present experiments using our proposed MMHOI dataset as well as the CORE4D dataset. For MMHOI, we use 16.2k images for training and 4.1k for testing. In order to evaluate the generalization of our approach to other datasets, we provide experiments using the recently introduced CORE4D dataset, that proposes the task of collaborative human-object motion forecasting and interaction synthesis in cases where two humans interact with a single object. We present comparisons on CORE4D using repurposed prior approaches. We used 482 of their sequences for training and remaining for testing our approach.}

\vspace{-5pt}
\paragraph{Evaluation Metrics.}
We evaluate our framework on 3D reconstruction, action recognition, and interactive body part detection. For 3D reconstruction, we compute Chamfer Distance (CD) and Vertex-to-Vertex (V2V) distance to measure the accuracy of human and object mesh predictions. For action recognition and interactive body part detection, we report the average classification accuracy across all action and body part classes. For fair comparisons in 3D reconstruction, we apply Procrustes alignment before computing these metrics.
\D{In our evaluation, the `S'-variant applies Procrustes alignment independently to \emph{each} human-object pair, leading to large errors due to mismatches in multi-HOI scenes. In contrast, the `M'-variant uses global alignment, yielding more accurate spatial assignment of humans to their interacting objects.}

\vspace{-10pt}
\paragraph{Implementation Details.}
Our framework is trained on the MMHOI dataset using a ViT-L backbone, with pre-trained weights from Multi-HMR \cite{multi-hmr2024} for the Human Perception Head. The training process takes three days on a single NVIDIA RTX A6000 (48GB) GPU. While MMHOI provides ground truth annotations in both SMPL-H and SMPL-X formats, we use SMPL-X in our experiments. 

\begin{table}[t]
\begin{center}
\setlength\tabcolsep{5pt} 
\caption{Ablation studies assessing the component contributions in reconstruction and action recognition.}
\vspace{-8pt}
\label{table:ablation}
\begin{tabular}{|l|c|c|}
\hline
Reconstruction& \makecell{M Hum \\ CD $\downarrow$} & \makecell{M Obj \\ CD $\downarrow$} \\
\hline
HPH only & 9.35 & 30.25 \\
OPH (main patch only) & 8.92 & 26.75 \\
\D{OPH (full: main + sub-patches) }& 8.10& 23.74 \\
OPH (full) + $\mathcal{L}_{act}$  & 7.31& 22.97 \\
OPH (full) + $\mathcal{L}_{act}$ + $\mathcal{L}_{bp}$ & 7.29& 22.81 \\
OPH (full) + $\mathcal{L}_{act}$ + $\mathcal{L}_{bp}$ + $\mathcal{L}_{cons}$ &\textbf{6.47} & \textbf{21.02} \\
\hline
\hline
Action Recognition & \multicolumn{2}{c|}{Accuracy (\%) $\uparrow$} \\
\hline
w/o body part detection & \multicolumn{2}{c|}{59.99}\\
w/ body part detection &\multicolumn{2}{c|}{ \textbf{60.96}} \\
\hline 

\end{tabular}
\centering

\end{center}
\vspace{-15pt}
\end{table}

\begin{figure}[t]
\begin{center}
\includegraphics[clip, trim=0cm 0cm 0cm 0cm, width=1\linewidth]{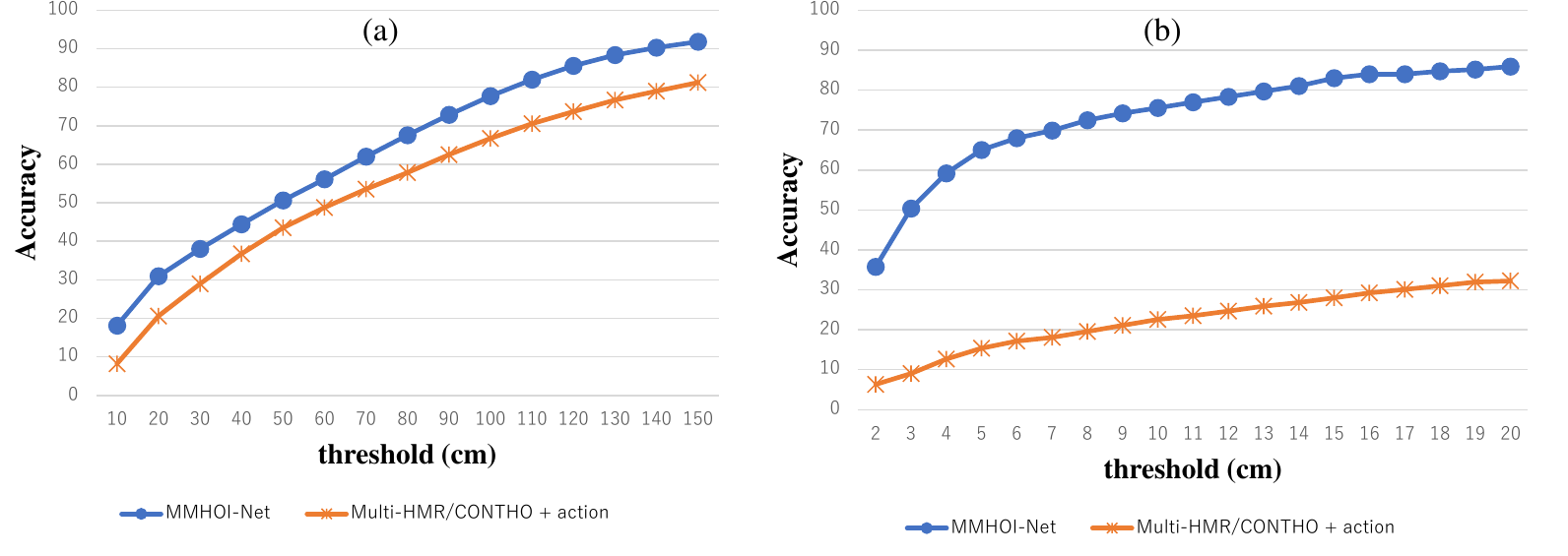}
\end{center}
\vspace{-13pt}
   \caption{
      \D{Evaluation of interaction prediction on MMHOI. (a) plots the \% of scenes where the predicted interaction body parts are close to the objects within a threshold (x-axis, in cm). (b) shows object-object interaction prediction comparison. Both plots evaluate multi-HOI accuracies after Procrustes alignment. The plots highlight the benefit of explicitly modeling multi-HOIs.}}
\label{fig:adde}
\vspace*{-10pt}
\end{figure}

\begin{figure*}[t]
\begin{center}
\includegraphics[clip, trim=0cm 0cm 0cm 0cm, width=1\linewidth]{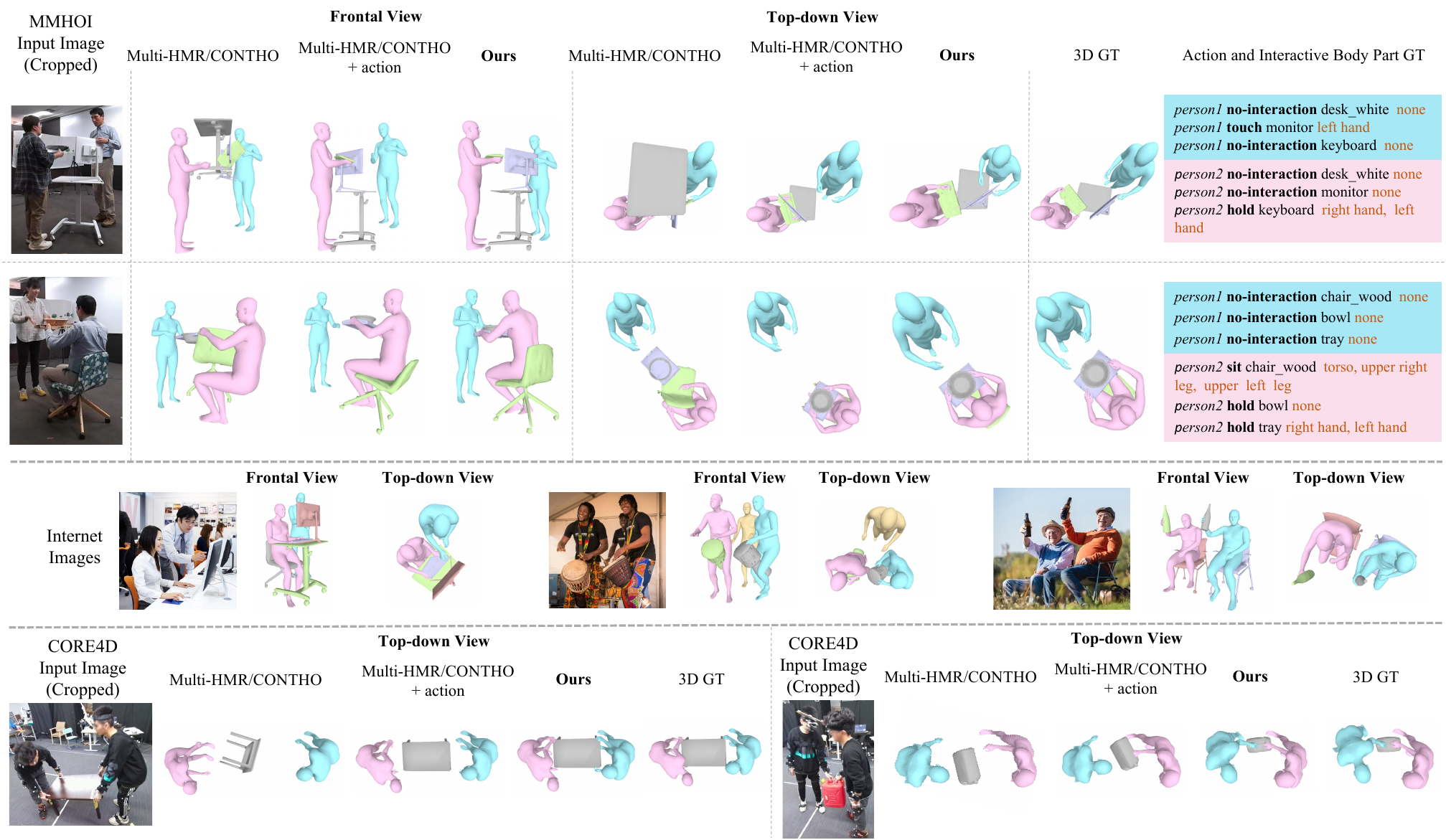}
\end{center}
\vspace{-10pt}
\caption{Qualitative comparisons of our method against Multi-HMR/CONTHO without and with action supervision \D{on MMHOI and CORE4D (S1) test sets.} Our approach shows better quality in human-object interaction prediction,  occlusion handling, and spatial consistency. The third row shows the zero-shot inference on Internet images demonstrating the generalization capabilities of our model. }
\label{fig:result_qualitative}
\end{figure*}

\vspace{-10pt}
\paragraph{State-of-the-art Comparisons.}
We compare our approach against state-of-the-art single-HOI reconstruction methods, including PHOSA \cite{zhang2020phosa}, CHORE \cite{xie2022chore}, and CONTHO \cite{nam2024joint}.\footnote{We do not include PICO \cite{cseke_tripathi_2025_pico} in our evaluation as it heavily relies on its own PICO-db dataset, which does not share object categories with ours.}
Since MMHOI focuses on multi-HOI scenarios where hand-object interactions are critical, we adopt SMPL-X-based Multi-HMR \cite{multi-hmr2024} for improved representation.
To ensure a fair comparison, we adapt PHOSA, CHORE, and CONTHO to work with Multi-HMR outputs by treating multi-HOI as independent single-HOI instances and aggregating their results. Specifically, human meshes are estimated using Multi-HMR’s SMPL-X outputs, while object poses are inferred separately for each individual in multi-human scenes. Since these methods are designed for single-HOI, we compute object meshes for shared interactions and obtain the final pose by averaging their predicted rotations \cite{rotation} and translations.
Table \ref{table:total_accuracy} shows that our method consistently outperforms these baselines, demonstrating improved multi-HOI reconstruction accuracy. 
Specifically, for Multi-HMR/PHOSA, Multi-HMR/CHORE, and Multi-HMR/CONTHO, we apply pairwise Procrustes alignment between humans and objects   (1--3 rows in Table \ref{table:total_accuracy}), as these methods were originally designed for single-HOI scenarios.
We also evaluate these methods with action supervision by integrating our action head (\secref{sec:hoi_head}), which incorporates multi-HOIs information. In this case, we perform evaluations on both single-HOI and multi-HOI settings (4--6 rows in Table \ref{table:total_accuracy}).
Although action supervision enhances the performance of past methods, our approach consistently surpasses them. 

\D{In the lower part of Table~\ref{table:total_accuracy}, we compare MMHOI-Net with recent reconstruction methods on the CORE4D \cite{liu2025core4d} seen-object (S1) test set, where our method achieves superior performance. We also assess interaction accuracy by measuring the alignment between predicted interactive body parts and object regions in Fig.\ref{fig:adde}(a), and show object-object interaction results in Fig.\ref{fig:adde}(b).}
These findings highlight the limitations of single-HOI methods, which fail to capture multi-HOIs dependencies, leading to suboptimal reconstructions in dense human-object interaction scenarios. We do not include HOI-$\mathrm{M}^3$ \cite{zhang2024hoi} in our evaluation as their full dataset and pre-trained models are unavailable.

\vspace{-10pt}
\paragraph{Qualitative Evaluations.}
\figref{fig:result_qualitative} presents qualitative comparisons between our method and Multi-HMR/CONTHO (with and without action supervision) across various object types, occlusions, and challenging viewpoints \D{on MMHOI and CORE4D datasets}. Our approach improves interaction reasoning and better preserves object affordance constraints, resulting in more coherent and semantically meaningful reconstructions. Additionally, we demonstrate the generalizability of our method by evaluating it on internet images without additional fine-tuning.

\vspace{-10pt}
\paragraph{Ablation Studies.}
Table \ref{table:ablation} shows that our full model performs best, while removing key components degrades accuracy, highlighting their importance in multi-HOI modeling. Specifically, we see that our dual-patch object representation leads to a 3 point drop in multi-object chamfer distance  against a `main patch only' representation. We also analyze the changes in action recognition performance when the body-parts are detected with +1\% improvement. See the supp. materials for more results, including results showing generalization to new objects.

\section{Conclusions}
We introduced MMHOI, the first large-scale dataset capturing the complexity of real-world \D{multi-human and multi-object-object} interactions, complete with 3D annotations, action labels, and interaction body parts. Building on MMHOI, we proposed MMHOI-Net, a transformer framework that reconstructs human-object geometries while reasoning about their interactions and actions. Experiments show MMHOI-Net outperforms prior methods in reconstruction accuracy and interaction consistency \D{on MMHOI and CORE4D datasets.} 

Yet, challenges remain -- severe occlusions disrupt predictions. Also, while we consider cooperative tasks, extending to full group interactions will require obtaining 3D ground truth of all interacting individuals, which we plan to explore in future work. Albeit these shortcomings,
we believe our work paves the way for future research on \D{3D multi-HOI}, providing a foundation for advancing HOI understanding. See the supp. materials for dataset samples, training details, and qualitative results.


{
    \small
    \bibliographystyle{ieeenat_fullname}
    \bibliography{mmhoi}
}

\twocolumn[
  \centering
  {\fontsize{13.5pt}{16pt}\selectfont \bf MMHOI: Modeling Complex 3D Multi-Human Multi-Object Interactions}\\
  \vspace{0.8em}
  {\fontsize{13.5pt}{16pt}\selectfont Supplementary Materials}
  \vspace{1.2em}
]
\setcounter{section}{0}
\renewcommand{\thesection}{\Alph{section}}
\setcounter{figure}{0} 
\renewcommand{\thefigure}{\thesection.\arabic{figure}}
\setcounter{table}{0} 
\renewcommand{\thetable}{\thesection.\arabic{table}}

This supplementary material provides details of MMHOI dataset construction introduced in Section 3, details of the method in Section 4, additional results of MMHOI-Net in Section 5, also failure cases of our proposed method.

\vspace{-5pt}
\section{Details of MMHOI Dataset Construction}
\paragraph{Dataset Statistic.}
\begin{figure}[t]
\begin{center}
\includegraphics[clip, trim=10cm 0cm 11cm 0cm, width=1\linewidth]{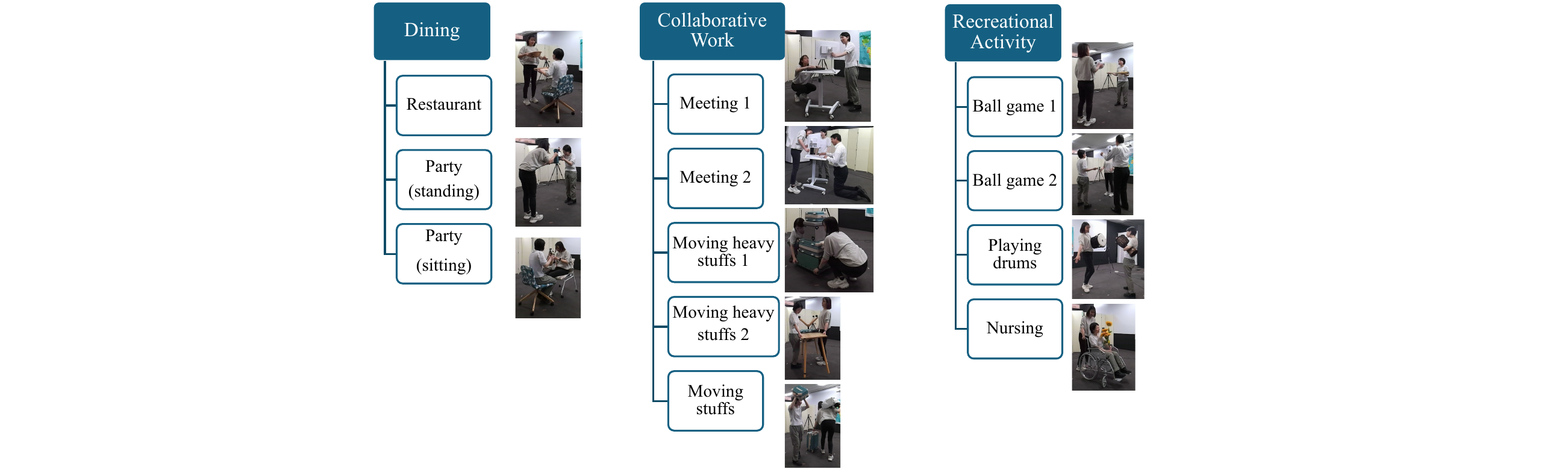}
\end{center}
\vspace{-5pt}
   \caption{The MMHOI dataset is categorized into three main interaction types: dining, collaborative work, and recreational activities. These interaction types span 12 daily-life scenarios.}
\label{fig:12}
\end{figure}

\begin{figure}[h]
\begin{center}
\includegraphics[clip, trim=0cm 0cm 0cm 0cm, width=1\linewidth]
{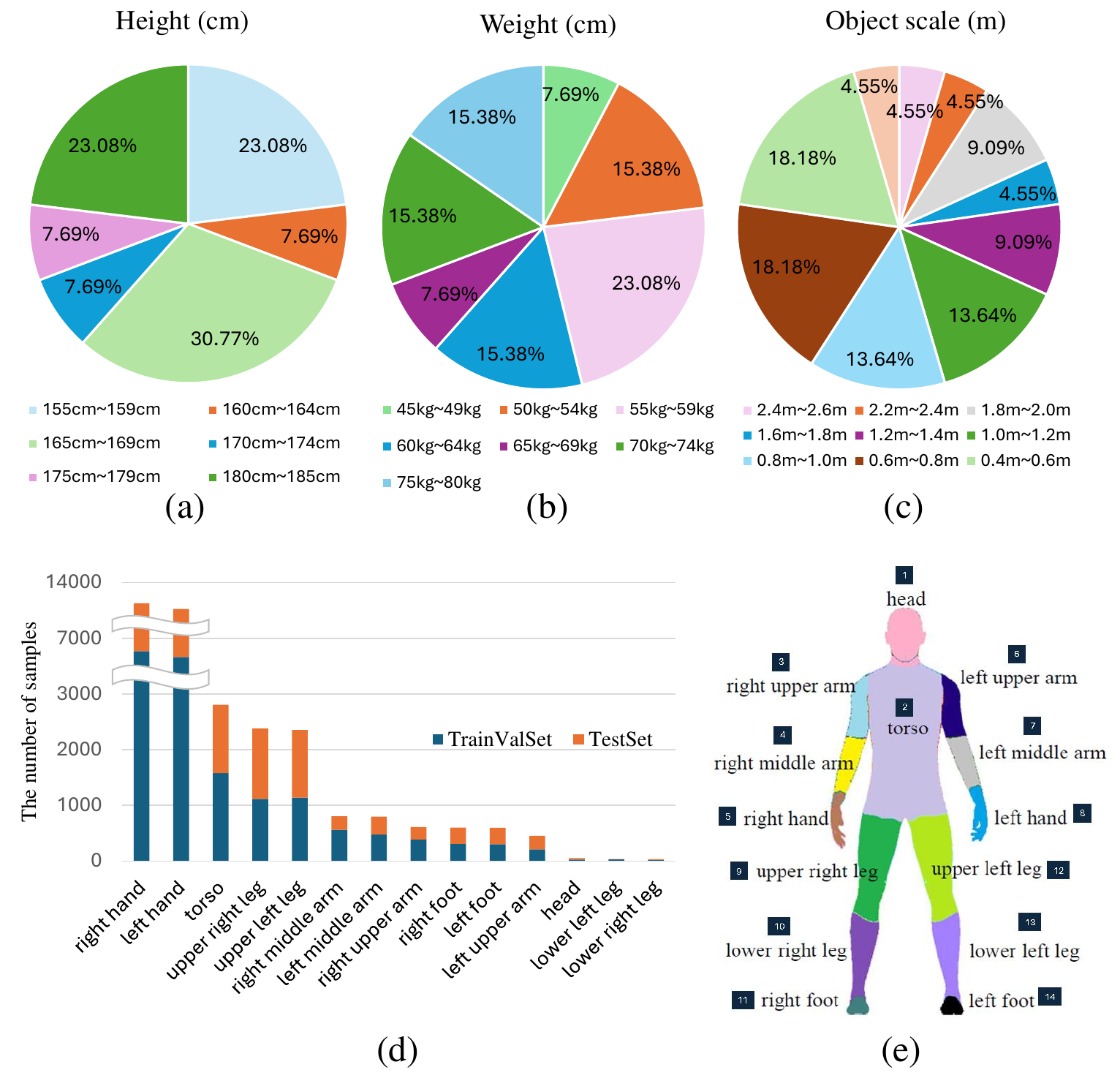}
\end{center}
\vspace{-5pt}
\caption{Statistics of humans and objects in the MMHOI dataset. (a) actor height, (b) actor weight, (c) object scales, (d) sample size distribution of interactive body part classes, and (e) 14 predefined body parts.}
\label{fig:sta}
\end{figure}

As shown in \figref{fig:12}, MMHOI dataset is categorized into three main interaction types -- dining (3 scenarios), collaborative work (5 scenarios), and recreational activities (4 scenarios) -- spanning 12 daily-living scenarios to ensure a balanced distribution of interactions.
\figref{fig:sta} and \figref{fig:sta2} present the statistical distributions of actors, objects, and interactions. \figref{fig:sta}(a)-(d) depict actor height, weight, object dimensions (length, width, height), sample distribution of interactive body part classes, while \figref{fig:sta}(e) illustrates the 14 predefined body parts. \figref{fig:sta2} shows the sample distribution of 78 action classes, with 13 rare classes containing fewer than 35 test samples.

\begin{figure*}[t]
\begin{center}
\includegraphics[clip, trim=0cm 0cm 0cm 0cm, width=1\linewidth]
{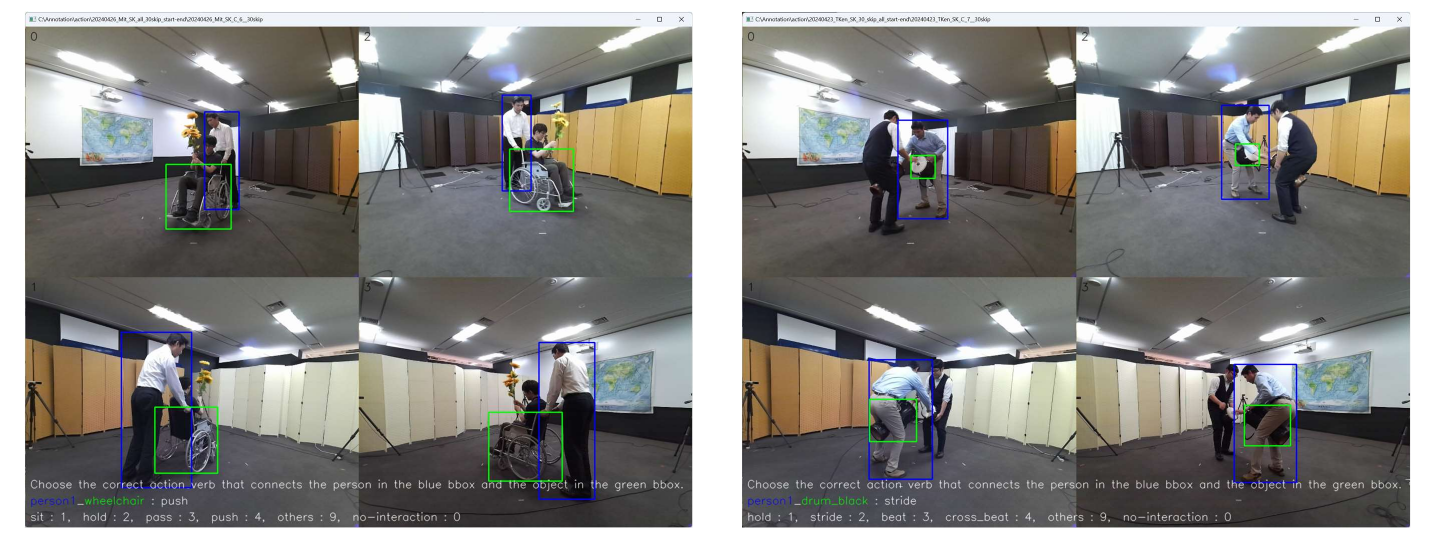}
\end{center}
\vspace{-5pt}
   \caption{Action annotation UI screen capture. Annotator is asked to choose the correct action verb's number from the presented candidate verbs.} 
\label{fig:anno}
\end{figure*}

\begin{figure}[t]
\begin{center}
\includegraphics[clip, trim=0cm 0cm 0cm 0cm, width=1\linewidth]{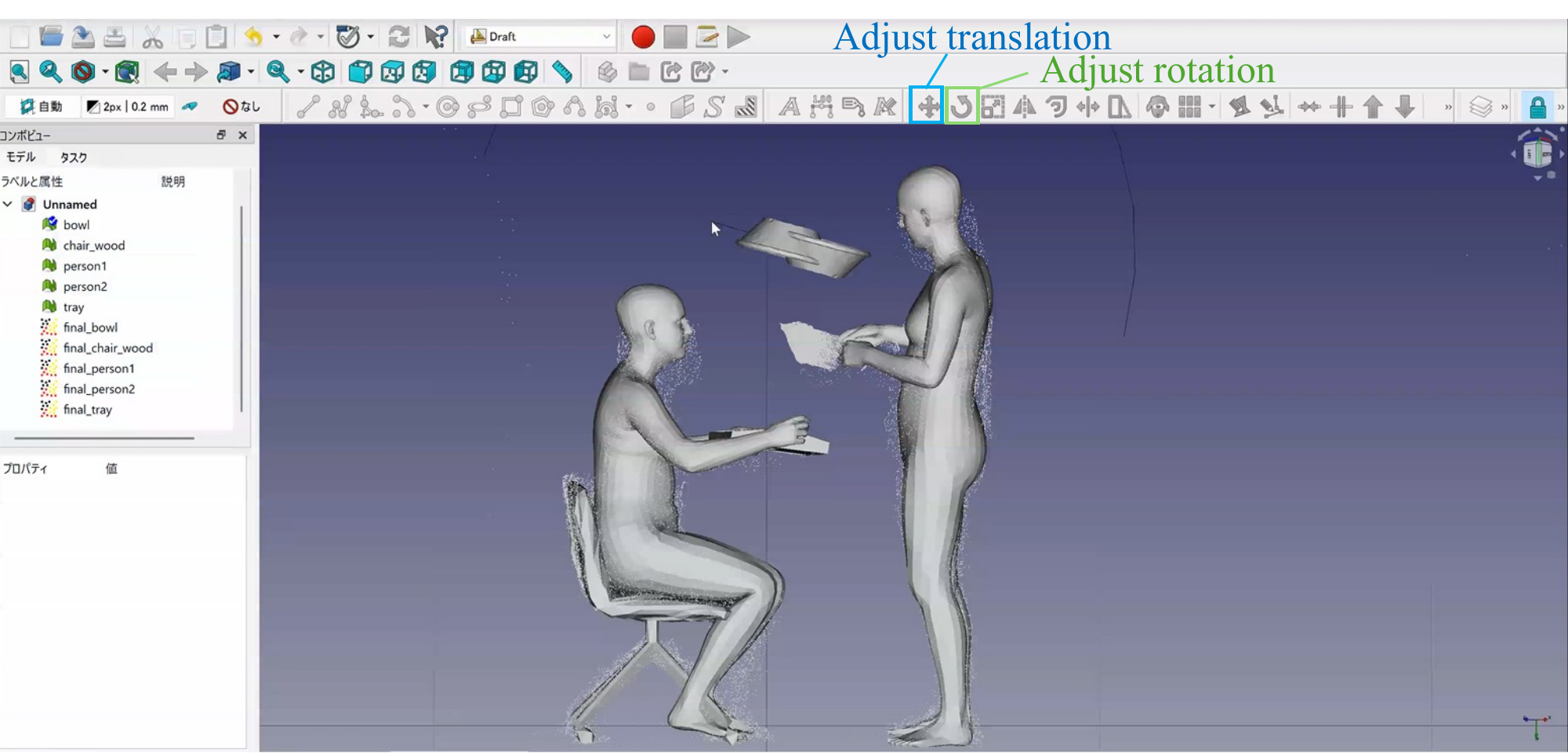}
\end{center}
\vspace{-3pt}
   \caption{CAD annotation screen capture. CAD operator adjust rotation and translation to finalize the 3D GT.}
\label{fig:cad}
\end{figure}

\vspace{-10pt}
\paragraph{Action Annotation.}
The action annotation interface is shown in \figref{fig:anno}, facilitates the labeling of human-object interactions within the dataset. Each action annotation is independently verified by at least two annotators to maintain accuracy. Annotators are responsible for assigning appropriate verb labels to each human-object pair, capturing the nature of their interaction.
The MMHOI dataset contains 78 human-object action types, with 21 unique action verbs. In contrast, CORE4D includes only 5 verb categories: ``move (together)'', ``raise (together)'', ``rotate (together)'', ``pass'', and ``others'', making MMHOI significantly more diverse in terms of action representation.

\vspace{-10pt}
\paragraph{3D Annotation.}
As shown in \figref{fig:cad}, 3D ground truth (GT) annotations are obtained via a two-step process. Each object’s location is initialized at the center of the four cameras.  
We first apply Iterative Closest Point (ICP) alignment with depth data for initial registration, followed by manual refinement in FreeCAD \cite{riegel2016freecad}. This refinement process involves adjusting human and object positions using translation and rotation tools (blue and green bounding boxes in \figref{fig:cad}).
Each annotation then undergoes verification by three CAD operators before being finalized.
The GT annotation records each object’s displacement and orientation relative to its initialized position.

\section{Details of Experiments}
\paragraph{Implementation Details.}
We initialize the MMHOI-Net weights using the multiHMR\_672\_S pre-trained model \cite{multi-hmr2024}, which processes images at a resolution of 672$\times$ 672. However, human meshes for baseline methods are inferred using the multiHMR\_896\_L model from the Multi-HMR GitHub repository \cite{multi-hmr2024}.
MMHOI-Net is trained on the MMHOI dataset with a batch size of 32 and a patch size of 24. 
During optimization, the loss weights are set as follows: $\lambda_h = 1$, $\lambda_{param} = 1$, $\lambda_{det} = 1$, $\lambda_{o} = 10$, $\lambda_{p} = 10$, $\lambda_{main} = 10$, $\lambda_{sub} = 10$, $\lambda_{act} = 100$, $\lambda_{bp} = 10$, and $\lambda_{cons} = 100$. Since we initialize human-related parameters using Multi-HMR \cite{multi-hmr2024}, the corresponding loss weights ($\lambda_{h}$, $\lambda_{param}$, $\lambda_{det}$) are set lower relative to the others.
We train MMHOI-Net using automatic mixed precision \cite{micikevicius2018mixed} for 500k iterations. Following \cite{multi-hmr2024}, we adopt SMPL-X with 10 shape components during both training and inference.

\begin{figure}[t]
\begin{center}
\includegraphics[clip, trim=2cm 0cm 2cm 0cm, width=1\linewidth]{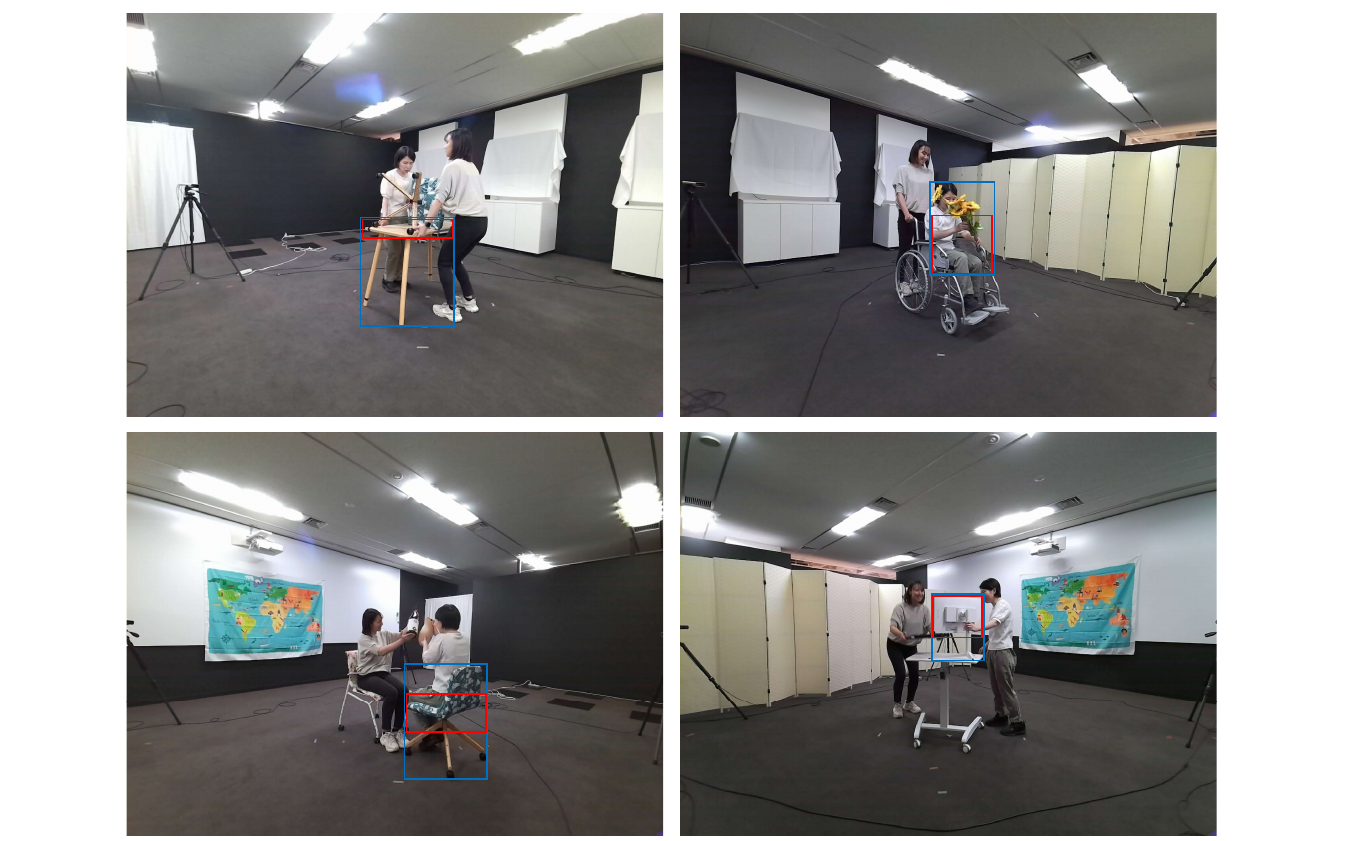}
\end{center}
\vspace{-8pt}
   \caption{Patch mask region per object. The blue bounding box indicates the whole L-SAM detected region while red bounding box indicates mask region used for the main and sub patches offset.}
\label{fig:patch}
\end{figure}

\vspace{-15pt}
\paragraph{Object Patch Selection and Mask Refinement.}
To ensure precise object representation, we refine the L-SAM masks used for defining main and sub patches, particularly for large objects such as tables, chairs, monitors, and flowers. As shown in \figref{fig:patch}, we apply mask shrinking strategies to 7 out of the 22 object categories to focus on the most interaction-relevant regions.
For chair objects, the mask is restricted to the bounding box height range of 20\%-55\% from the top. Similarly, for tables, monitors, and flowers, the mask is limited to $0\%-15\%$, $0\%-60\%$, $40\%-100\%$ of the bounding box height from the top, respectively.

\vspace{-10pt}
\paragraph{\D{Effective Dataset Size}} 
Our dataset has 4--14 unique action classes per scenario. Further, our training and test sets are balanced across 12 scenarios, each with $\sim$50K frames. During training, we used a validation set to avoid overfitting.

\section{Additional Results and Details}

\begin{figure*}[t]
\begin{center}
\includegraphics[clip, trim=0cm 0cm 0cm 0cm, width=1\linewidth]
{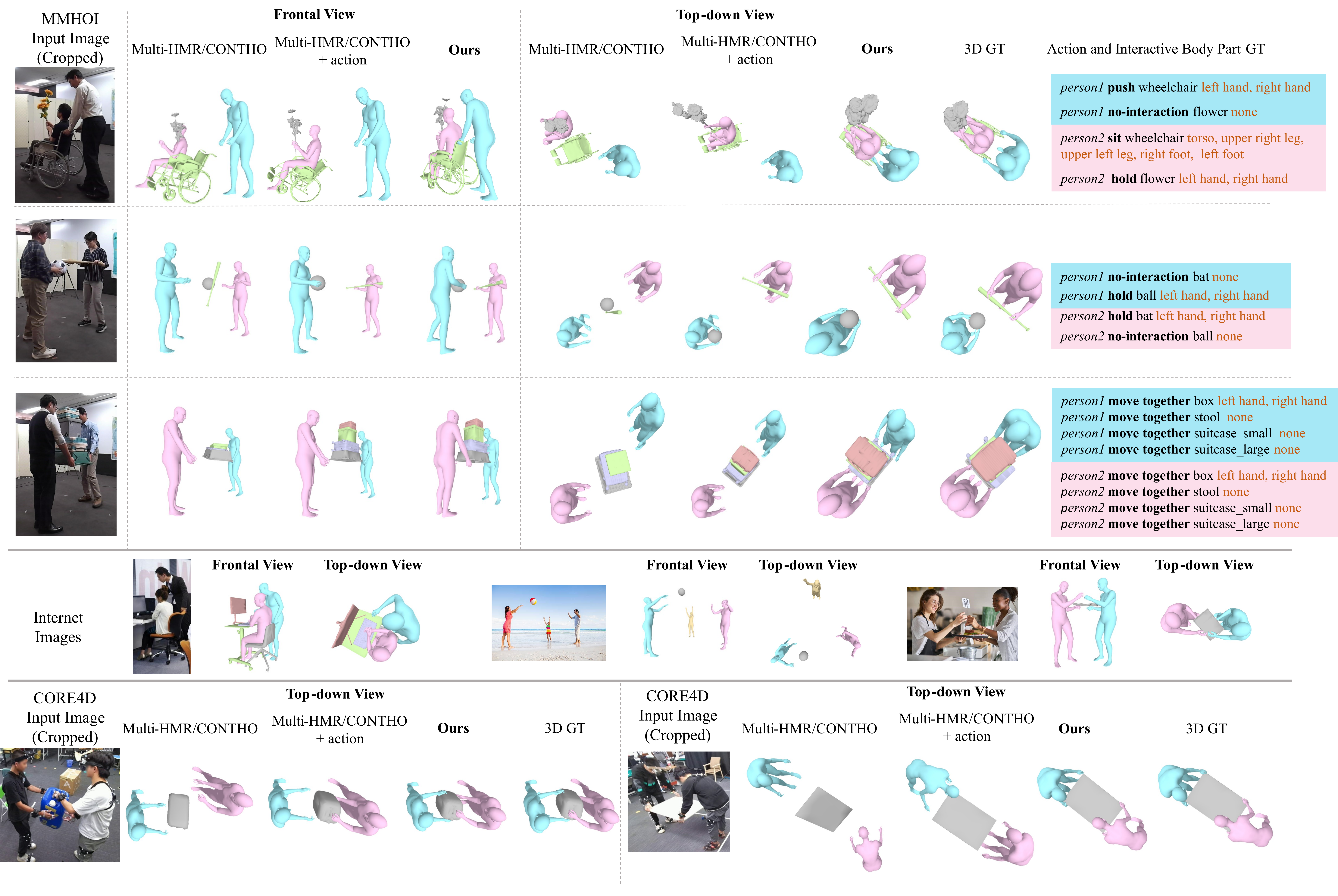}
\end{center}
\caption{Additional results comparing MMHOI-Net with Multi-HMR/CONTHO, with and without action supervision, \D{on MMHOI and CORE4D (S1) test sets}. Our method outperforms prior approaches in inferring reasonable multi-HOI reconstructions.}
\label{fig:mmhoi}
\end{figure*}

\begin{figure*}[t]
\begin{center}
\includegraphics[clip, trim=0cm 0cm 0cm 0cm, width=1\linewidth]
{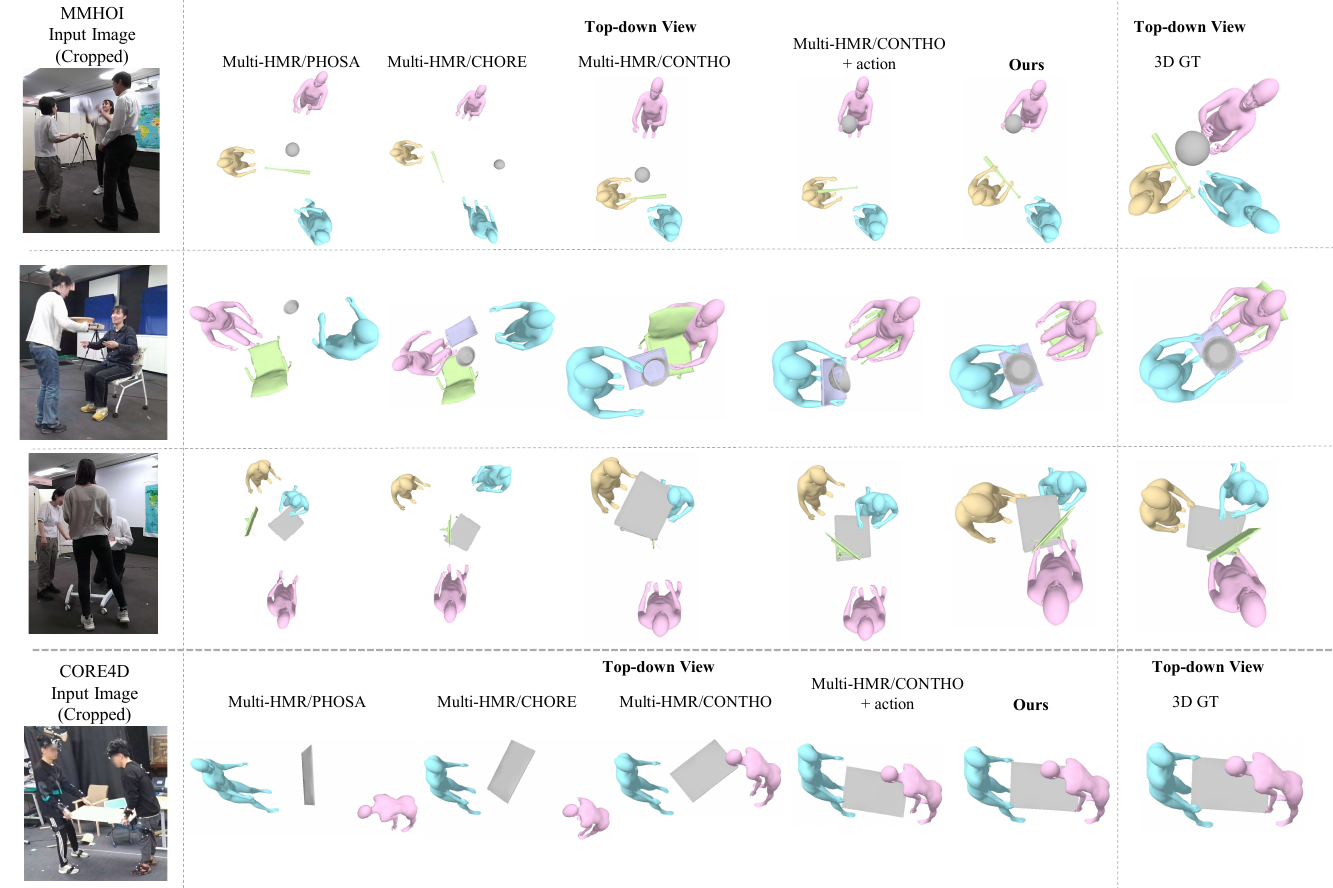}
\end{center}
\caption{Additional results comparing MMHOI-Net with Multi-HMR/CONTHO (with and without action supervision), Multi-HMR/CHORE and Multi-HMR/PHOSA \D{on MMHOI and CORE4D (S1) test sets.} Our method outperforms past methods in multi-HOI 3D reconstructions.}
\label{fig:chore}
\end{figure*}

\begin{figure}
\begin{center}
\includegraphics[width=1.0\linewidth]{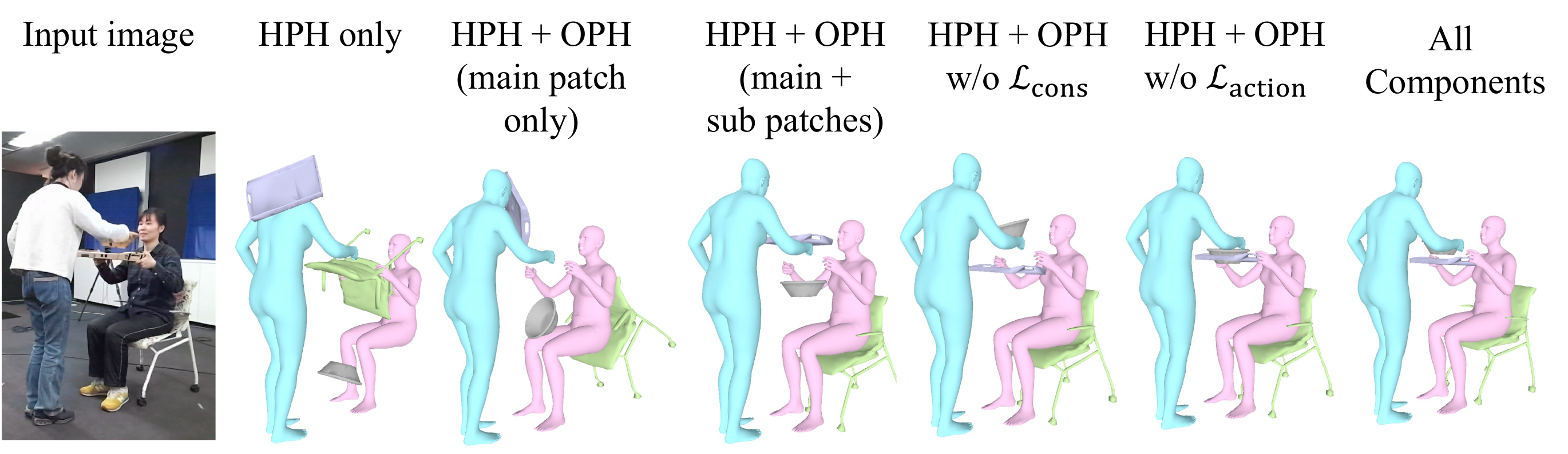}
\end{center}
\caption{\D{Additional ablation studies. Our proposed dual-patch representation approach and losses improve reconstruction accuracy and interaction consistency.}
}
\label{fig:with}
\end{figure}

\begin{figure}[t]
\begin{center}
\includegraphics[clip, trim=0cm 0cm 0cm 0cm, width=1\linewidth]{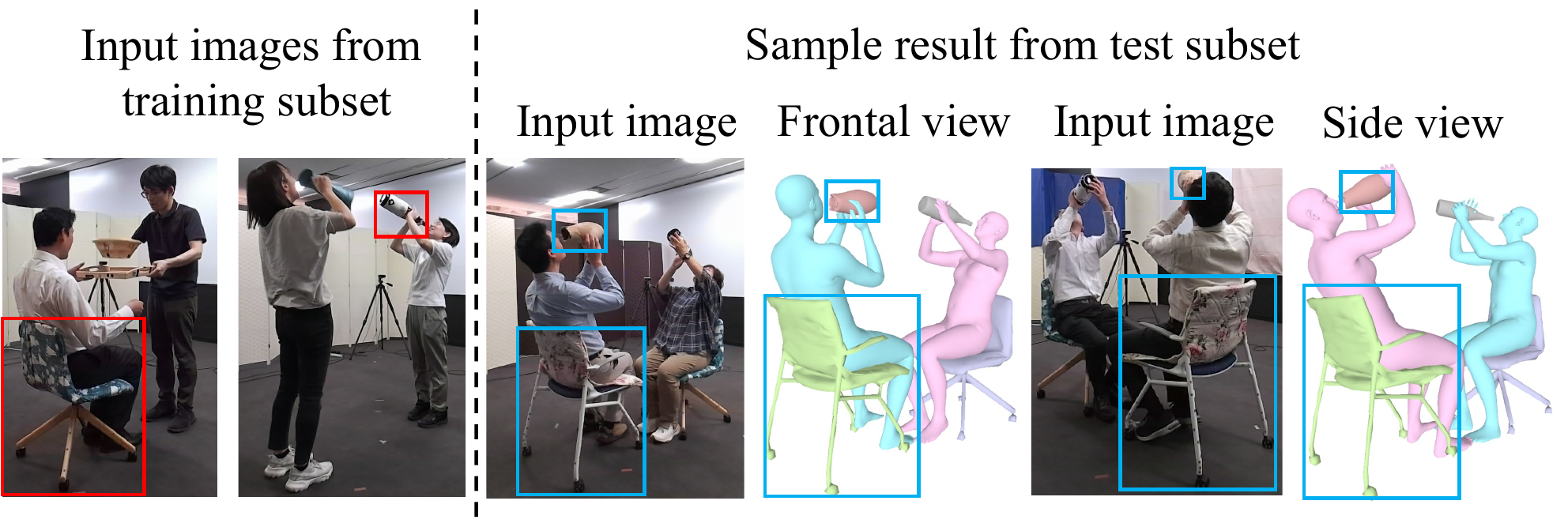}
\end{center}
\caption{\D{Generalization results to objects unseen during training on MMHOI dataset. Red and blue boxes indicate representative seen and unseen objects respectively.}}
\label{fig:generalization}
\end{figure}

\paragraph{Additional Visual Examples}
\figref{fig:mmhoi} presents additional qualitative comparisons between our method and Multi-HMR/CONTHO, including results with and without action supervision, as well as in-the-wild internet images on MMHOI \D{and CORE4D (S1) test sets.} \figref{fig:chore} shows samples of comparing our method with Multi-HMR/PHOSA, Multi-HMR/CHORE and Multi-HMR/CONTHO (with and without action supervision) on MMHOI and CORE4D (S1) test sets. Our model, trained on MMHOI, demonstrates superior reconstruction of multi-human-object interactions (multi-HOIs) compared to prior methods. Notably, our approach achieves more plausible and coherent reconstructions, effectively capturing human-object affordances and reducing physically implausible configurations.

\vspace{-12pt}
\paragraph{Additional Ablation Studies.}
\D{\figref{fig:with} demonstrates the impact of our dual-patch representation approach and losses, showing its effectiveness in enhancing 3D reconstruction accuracy and enforcing HOI consistency. \tabref{table:action_accuracy} reports action recognition and interactive body part detection performance across various human-object interactions on MMHOI and CORE4D (S1). 
\D{CORE4D does not provide body part annotations, so we generate them as follows.
For each body part, we compute the Chamfer Distance $\Psi$ between its point set $\mathcal{P}^i_\mathrm{bp}$ and the corresponding object point set $\mathcal{P}_\mathrm{o}$. If $\Psi < \delta$ (with threshold $\delta = 5mm$), we consider the body part to be interacting with the object; otherwise, no interaction is assigned.}
}

\begin{table}[t]
\begin{center}
\setlength\tabcolsep{3.5pt} 
\caption{Action recognition and interactive body part detection accuracy (\%) on MMHOI-Net. For MMHOI dataset, by ``Rare'', we mean 13 action classes that have fewer than 35 test samples, while ``Non-Rare'' includes the remaining action categories. ``All'' represents the overall accuracy across all action categories on MMHOI and CORE4D (S1) test set. 
}
\label{table:action_accuracy}
\begin{tabular}{|l|c|c|c|c|c|}
\hline
Dataset & Rare & Non-Rare & All & Body Part\\
\hline\hline
MMHOI&  12.1 & 70.3 &61.0 &67.0\\
\hline
\hline
CORE4D (S1)&  - & - &76.5 &88.7\\
\hline
\end{tabular}
\end{center}
\end{table}

\vspace{-12pt}
\paragraph{Generalization to New Objects}
To analyze generalization of our method to new objects, we re-trained our MMHOI-Net without 5 object categories (plastic chair, mug, drum brown, backpack, small suitcase) on MMHOI dataset and tested it to predict the poses of these objects (\figref{fig:generalization}). 
\D{We also evaluate MMHOI-Net on CORE4D’s unseen object test set (S2). CORE4D provides two test sets: seen objects (S1) and unseen objects (S2). Results for S1 are presented in Table 2 and Figure 6 of the main paper. As shown in \tabref{table:gene}, our method outperforms prior approach on the S2 split.}
We believe the action cues allow our model to generalize and predict plausible HOI dynamics for unseen objects (of the same affordance).

\begin{table}[t]
\begin{center}
\caption{Generalization performance using 5 unseen objects on MMHOI \D{test subset, unseen objects test set on CORE4D (S2).}}
\label{table:gene}
\resizebox{0.48\textwidth}{!}{\begin{tabular}{|c|c|c|c|}
\hline
Dataset & Method& \makecell{S Human \\ Chamfer $\downarrow$} & \makecell{S Object \\ Chamfer $\downarrow$} \\
\hline
 MMHOI &Multi-HMR + CONTHO& 7.47 &  87.16 \\
&MMHOI-Net &\textbf{6.59} & \textbf{64.51} \\
\hline
\hline
 CORE4D &Multi-HMR + CONTHO& 23.93 &  198.49 \\
(S2)&MMHOI-Net &\textbf{19.31} & \textbf{151.82} \\
\hline
\end{tabular}
}
\centering
\end{center}
\end{table}

\section{Details of Interaction Evaluation (Fig. 5 main paper)}
\begin{figure}[ht]
\begin{center}
\includegraphics[width=1.0\columnwidth]{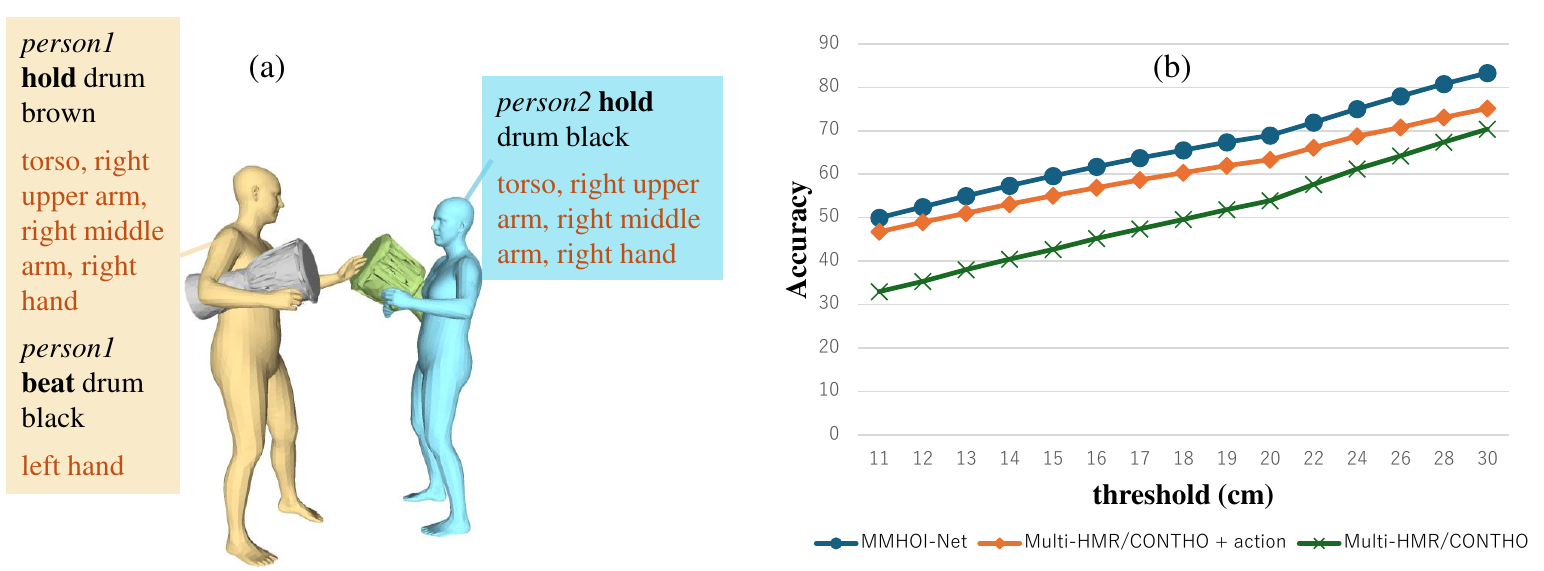}
\end{center}
\caption{
\D{(a) Example to illustrate the evaluation of interactions described in Fig. 5(a) of the main paper, and (b) of this figure. (b) shows evaluation of Single-HOI after Procrustes alignment.}
}
\label{fig:eval}
\end{figure}
\D{In this section, we detail the evaluation scheme used to compute the accuracy metrics presented in Fig. 5(a) and 5(b) of the main paper. For clarity, we illustrate the process using the setting shown in \figref{fig:eval}(a). Specifically, we evaluate interaction prediction in the Single HOI setting (\figref{fig:eval}(b)), Multi-HOI setting (Fig. 5(a) main paper), and object-object interactions (Fig. 5(b) main paper).
}

\vspace{-10pt}
\D{\paragraph{Single Body Part-Object Interaction Evaluation:} We compute the Chamfer Distance between the SMPL body part and object meshes after Procrustes alignment. In this case, our goal is to predict the alignment of our prediction (source) with the transformed target (GT after Procrustes alignment) for each potential interaction separately. To this end, we compute the individual pairwise interactions, compute the alignment with GT, and average over the matches. For example, with respect to Fig.~\ref{fig:eval} (a), as seen there is interaction between the SMPL model of the yellow person and the drum brown. We will consider predicted regions for the torso, right upper arm, right middle arm, and right hand for the yellow person and compute their distances to the drum brown mesh, and check if each of these distance are less than the given threshold against the corresponding aligned GT meshes. We repeat this for all the interaction regions for the entire scene and average over them to compute the accuracy at a given threshold. We vary this threshold to produce the left plot in \figref{fig:eval} (b).  }

\vspace{-15pt}
\D{\paragraph{Multiple Body Part-Object Interaction Evaluation:} 
We compute the result in (a) of Fig. 5 as follows. We compute all points of contact body parts between a person and an object, and instead of averaging over each HOI separately, we compute the entire set of interactions as a whole, by adding up the misalignments with the GT meshes. If the total misalignment is more than a given threshold, we assume the interaction prediction is a failure. We repeat this process for varied thresholds to produce the plot in Fig. 5(a).  With respect to the Fig.~\ref{fig:eval}(a), in this case, we assume the model predicts all contact body part points of the person in yellow (torso, right upper arm, right middle arm, right hand, and left hand). We also compute all the contact body part points of the person in blue (torso, right upper arm, right middle arm, right hand). We next compute the objects in contact with each person (drum brown and drum black). We compute the Procrustes alignment of the GT points with all the objects to compute the chamfer distances, which are added together to find the full error.}

\vspace{-10pt}
\paragraph{Object-Object Interaction Evaluation:} 
\D{Fig. 5(b) in the main paper presents performance on object-object interaction prediction. To evaluate this, we extract a test subset from MMHOI containing all samples with object-object interactions.}
\D{We compute the result in Fig. 5(b) as follows. We apply Procrustes alignment for global matching in Multi-HOI meshes before computing the chamfer distance. We compute the alignment of the GT points from the object-object pair to compute the chamfer distance. By adding up the misalignments with the GT object meshes, if the total misalignment is more than a given threshold, we assume the object-object interaction prediction is a failure. We repeat this process for varied thresholds to produce the plot in Fig. 5(b). As shown in Fig. 5 (b), our MMHOI-Net outperforms a past method which highlighting our method implicitly capture multi-object-object poses from the novel object dual-patch representation.}

\section{Limitations in Details.}
\begin{figure}[t]
\begin{center}
\includegraphics[clip, trim=0cm 0cm 0cm 0cm, width=1\linewidth]{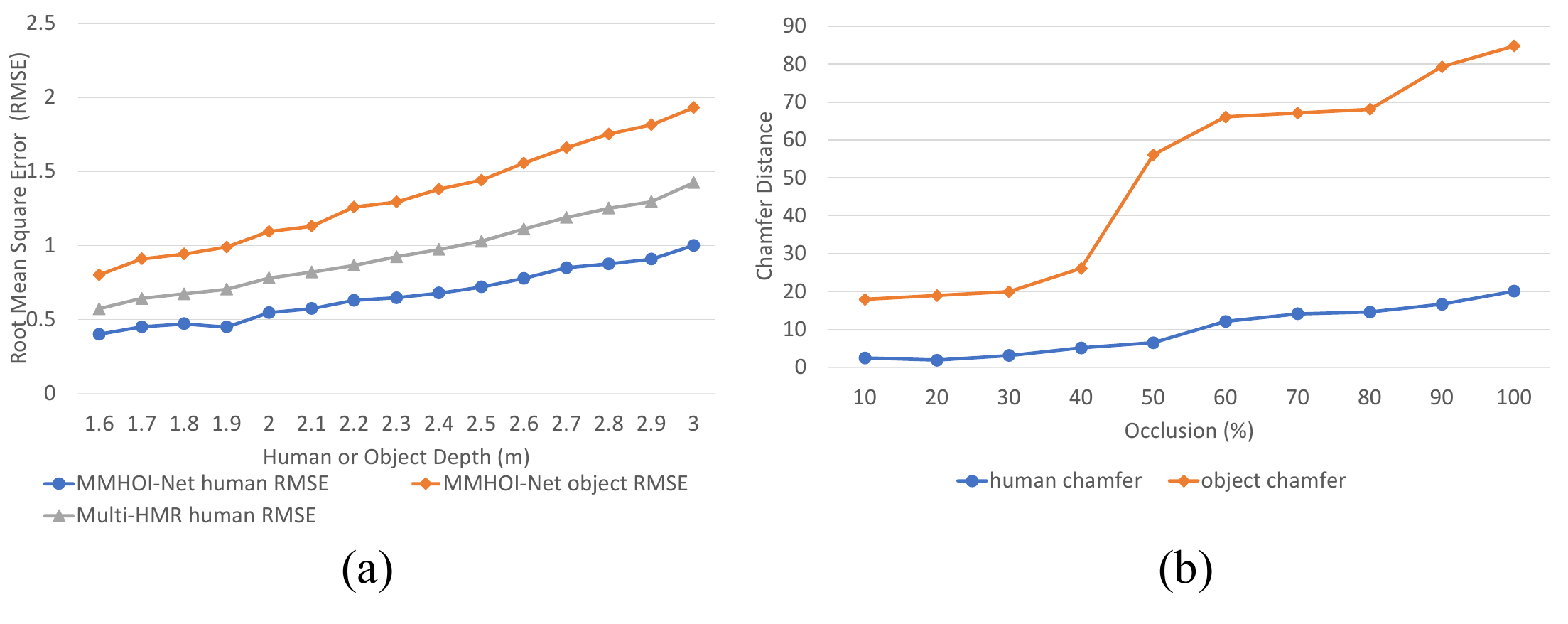}
\end{center}
\vspace{-10pt}
\caption{\D{In (a), we report the RMSE (m) per depth on MMHOI-Net and Multi-HMR. In (b), we show Chamfer distance (cm) per occlusion rate (\%) on MMHOI-Net.}}
\label{fig:occlusion}
\end{figure}

\begin{figure}[t]
\begin{center}
\includegraphics[clip, trim=0cm 0cm 0cm 0cm, width=1\linewidth]{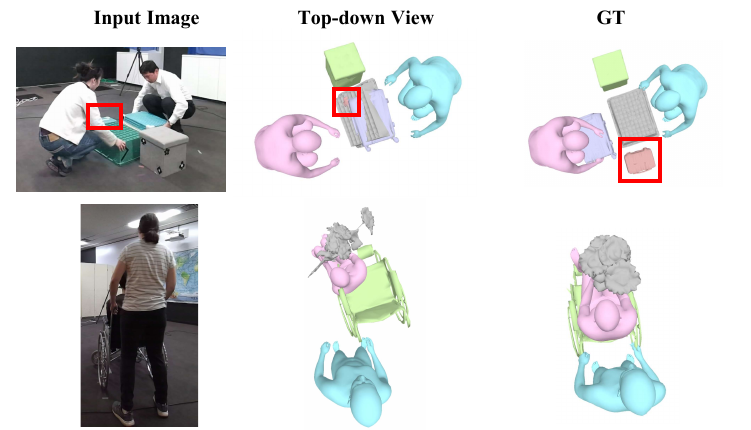}
\end{center}
\vspace{-15pt}
\caption{Failure cases of MMHOI-Net predictions. Severe occlusions and ambiguous ``no-interaction'' cases lead to inaccurate reconstructions.}
\label{fig:fail}
\end{figure}

\paragraph{Single-view and Occlusions}
Our model mitigates single-view ambiguity using our dual-patch representation and action cues. In \D{\figref{fig:occlusion} (a)}, we report RMSE in depth prediction (m) using pelvis as the root for human and the object center as root for the object. As past methods, e.g., CHORE, do not infer the absolute depth, we could compare ours only to Multi-HMR. \D{\figref{fig:occlusion} (b)} shows our method's effectiveness in handling partial occlusions. Severe occlusions of humans or objects remain challenging and often lead to inaccurate action predictions and reconstruction errors.
\figref{fig:fail} shows typical failure cases of MMHOI-Net. A potential improvement is leveraging temporal consistency \cite{xie2023vistracker} through multi-frame information.

\paragraph{Group Interactions}
Our N$\times$M pairwise interaction structure focuses on HOI pairs and may not capture the complexity of true group interactions. 
While we do consider cooperative tasks (e.g., ``object-move together''), extending to full group interactions will require obtaining 3D ground truth of all interacting individuals, which we plan to explore in future work.

\section{Analysis of CORE4D Dataset}
\begin{figure}[t]
\begin{center}
\includegraphics[clip, trim=0cm 0cm 0cm 0cm, width=1\linewidth]{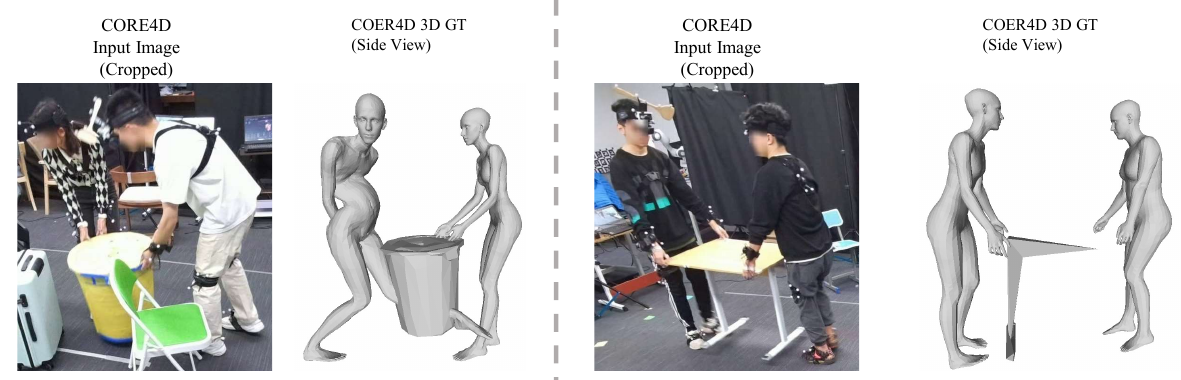}
\end{center}
\caption{CORE4D dataset have quality problems in SMPL-X model (left sample) and object mesh (right sample).}
\label{fig:core4d}
\end{figure}
\D{In Tab. 2 of the main paper, Chamfer Distance and V2V errors on the CORE4D test set are higher than those on the MMHOI dataset for both prior methods and MMHOI-Net. This is primarily due to the lower quality of 3D ground truth in CORE4D \cite{liu2025core4d}. As illustrated in \figref{fig:core4d} (left), the human mesh exhibits physically implausible SMPL-X configurations, while the object mesh (right) also appears unrealistic. Additionally, some sequences in CORE4D are not synchronized.}

\begin{figure*}[t]
\begin{center}
\includegraphics[clip, trim=0cm 0cm 0cm 0cm, width=0.87\linewidth]
{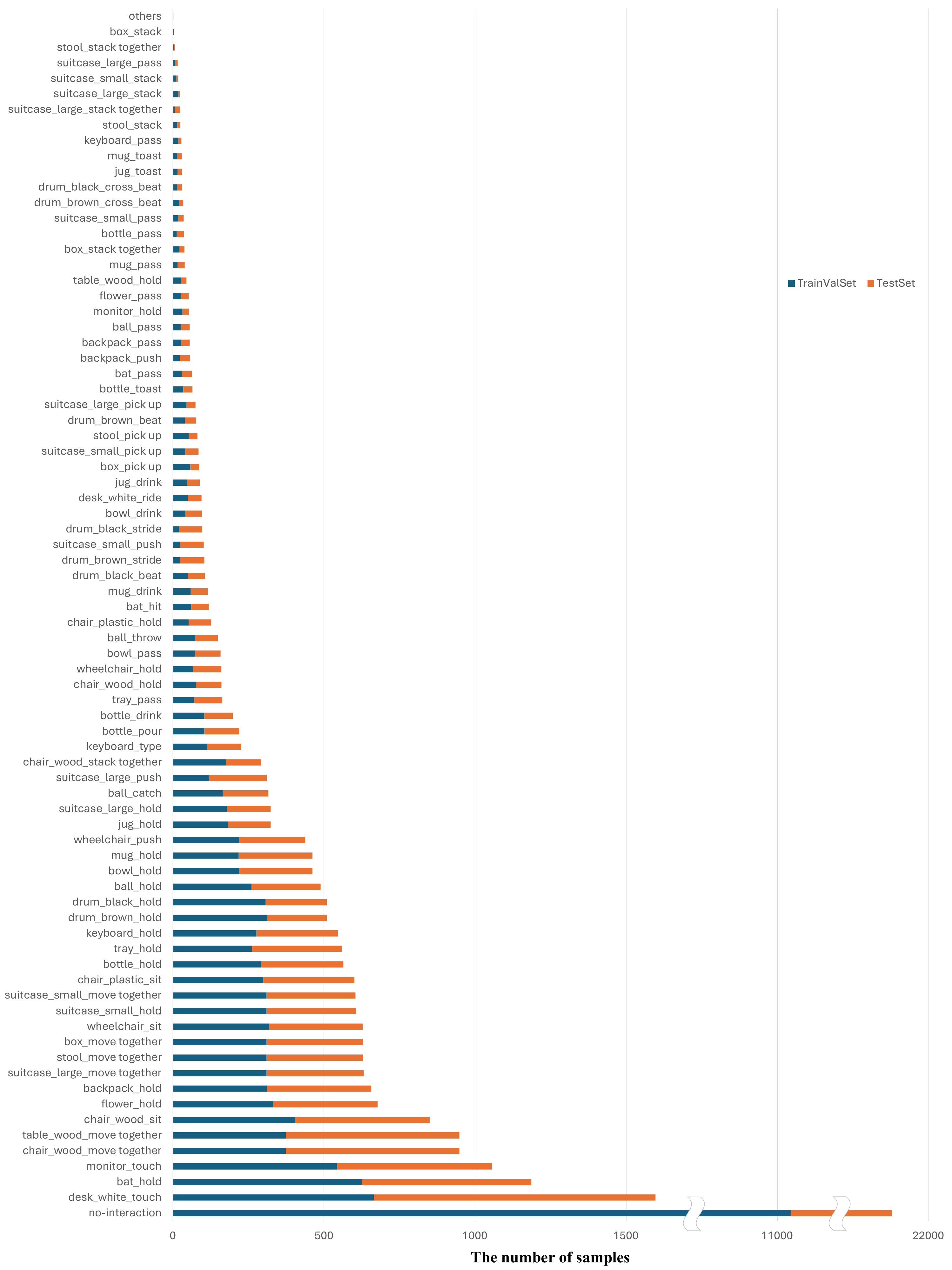}
\end{center}
\vspace{-15pt}
    \caption{Sample distribution of action classes in the MMHOI dataset.}
\label{fig:sta2}
\end{figure*}

\end{document}

%% file: preamble.tex
\usepackage{times}
\usepackage{epsfig}
\usepackage{graphicx}
\usepackage{amsmath}
\usepackage{amssymb}

\usepackage{caption}
\usepackage{cuted}
\usepackage{multirow}
\usepackage{makecell}
\usepackage{url}

\newcommand{\projectlink}[1]{%
    \href{#1}{{\small \textcolor[HTML]{FF69B4}{\rmfamily #1}}}%
}

\newcommand{\figref}[1]{Fig.\,\ref{#1}}

\newcommand{\secref}[1]{Sec.\,\ref{#1}}

\newcommand{\tabref}[1]{Tab. \,\ref{#1}}

\def\D#1{\textcolor{black}{#1}}

\def\AC#1{\textcolor{black}{#1}}

\def\eg{\emph{e.g}.} 


%% file: mmhoi.bib
@inproceedings{SMPL-X:2019,
  title = {Expressive Body Capture: {3D} Hands, Face, and Body from a Single Image},
  author = {Pavlakos, Georgios and Choutas, Vasileios and Ghorbani, Nima and Bolkart, Timo and Osman, Ahmed A. A. and Tzionas, Dimitrios and Black, Michael J.},
  booktitle = {Proceedings IEEE Conf. on Computer Vision and Pattern Recognition (CVPR)},
  pages     = {10975--10985},
  year = {2019}
}

@inproceedings{bhatnagar2022behave,
  title={{BEHAVE}: Dataset and method for tracking human object interactions},
  author={Bhatnagar, Bharat Lal and Xie, Xianghui and Petrov, Ilya A and Sminchisescu, Cristian and Theobalt, Christian and Pons-Moll, Gerard},
  booktitle={Proceedings of the IEEE/CVF Conference on Computer Vision and Pattern Recognition},
  pages={15935--15946},
  year={2022}
}

@inproceedings{cseke_tripathi_2025_pico,
    title     = {{PICO}: Reconstructing {3D} People In Contact with Objects}, 
    author    = {Cseke, Alp\'{a}r and Tripathi, Shashank and Dwivedi, Sai Kumar and Lakshmipathy, Arjun and Chatterjee, Agniv and Black, Michael J. and Tzionas, Dimitrios},
    booktitle = {Proceedings of the IEEE/CVF Conference on Computer Vision and Pattern Recognition (CVPR)},
    month     = {June},
    year      = {2025},
}

@inproceedings{xie2022chore,
  title={Chore: Contact, human and object reconstruction from a single rgb image},
  author={Xie, Xianghui and Bhatnagar, Bharat Lal and Pons-Moll, Gerard},
  booktitle={European Conference on Computer Vision},
  pages={125--145},
  year={2022},
  organization={Springer}
}

@inproceedings{nam2024joint,
  title={Joint Reconstruction of 3D Human and Object via Contact-Based Refinement Transformer},
  author={Nam, Hyeongjin and Jung, Daniel Sungho and Moon, Gyeongsik and Lee, Kyoung Mu},
  booktitle={Proceedings of the IEEE/CVF Conference on Computer Vision and Pattern Recognition},
  pages={10218--10227},
  year={2024}
}

@InProceedings{zhang2020phosa,
    title = {Perceiving 3D Human-Object Spatial Arrangements from a Single Image in the Wild},
    author = {Zhang, Jason Y. and Pepose, Sam and Joo, Hanbyul and Ramanan, Deva and Malik, Jitendra and Kanazawa, Angjoo},
    booktitle = {European Conference on Computer Vision (ECCV)},
    year = {2020},
}

@inproceedings{
      zhang2023neuraldome,
      title={NeuralDome: A Neural Modeling Pipeline on Multi-View Human-Object Interactions},
      author={Juze Zhang and Haimin Luo and Hongdi Yang and Xinru Xu and Qianyang Wu and Ye Shi and Jingyi Yu and Lan Xu and Jingya Wang},
      booktitle={CVPR},
      year={2023},
}

@inproceedings{GRAB:2020,
  title = {{GRAB}: A Dataset of Whole-Body Human Grasping of Objects},
  author = {Taheri, Omid and Ghorbani, Nima and Black, Michael J. and Tzionas, Dimitrios},
  booktitle = {European Conference on Computer Vision (ECCV)},
  year = {2020},
  url = {https://grab.is.tue.mpg.de}
}

@article{huang2024intercap, 
    title        = {{InterCap}: Joint Markerless {3D} Tracking of Humans and Objects in Interaction from Multi-view {RGB-D} Images}, 
    author       = {Huang, Yinghao and Taheri, Omid and Black, Michael J. and Tzionas, Dimitrios}, 
    journal      = {{International Journal of Computer Vision (IJCV)}}, 
    volume       = {},
    number       = {},
    pages        = {},
    doi          = {10.1007/s11263-024-01984-1},
    year         = {2024}
}

@inproceedings{jiang2023chairs,
  title={Full-Body Articulated Human-Object Interaction},
  author={Jiang, Nan and Liu, Tengyu and Cao, Zhexuan and Cui, Jieming and Chen, Yixin and Wang, He and Zhu, Yixin and Huang, Siyuan},
  booktitle={ICCV},
  year={2023}
}

@article{mandery2016unifying,
  title={Unifying representations and large-scale whole-body motion databases for studying human motion},
  author={Mandery, Christian and Terlemez, {\"O}mer and Do, Martin and Vahrenkamp, Nikolaus and Asfour, Tamim},
  journal={IEEE Transactions on Robotics},
  volume={32},
  number={4},
  pages={796--809},
  year={2016},
  publisher={IEEE}
}

@inproceedings{fan2023arctic,
  title = {{ARCTIC}: A Dataset for Dexterous Bimanual Hand-Object Manipulation},
  author = {Fan, Zicong and Taheri, Omid and Tzionas, Dimitrios and Kocabas, Muhammed and Kaufmann, Manuel and Black, Michael J. and Hilliges, Otmar},
  booktitle = {Proceedings IEEE Conference on Computer Vision and Pattern Recognition (CVPR)},
  year = {2023}
}

@misc{lv2024himonewbenchmarkfullbody,
        title={HIMO: A New Benchmark for Full-Body Human Interacting with Multiple Objects}, 
        author={Xintao Lv and Liang Xu and Yichao Yan and Xin Jin and Congsheng Xu and Shuwen Wu and Yifan Liu and Lincheng Li and Mengxiao Bi and Wenjun Zeng and Xiaokang Yang},
        year={2024},
        eprint={2407.12371},
        archivePrefix={arXiv},
        primaryClass={cs.CV},
        url={https://arxiv.org/abs/2407.12371}, 
  }

@inproceedings{zhang2024hoi,
  title={HOI-M3: Capture Multiple Humans and Objects Interaction within Contextual Environment},
  author={Zhang, Juze and Zhang, Jingyan and Song, Zining and Shi, Zhanhe and Zhao, Chengfeng and Shi, Ye and Yu, Jingyi and Xu, Lan and Wang, Jingya},
  booktitle={Proceedings of the IEEE/CVF Conference on Computer Vision and Pattern Recognition},
  pages={516--526},
  year={2024}
}

@misc{kim2024parahome,
        title={ParaHome: Parameterizing Everyday Home Activities Towards 3D Generative Modeling of Human-Object Interactions}, 
        author={Jeonghwan Kim and Jisoo Kim and Jeonghyeon Na and Hanbyul Joo},
        year={2024},
        eprint={2401.10232},
        archivePrefix={arXiv},
        primaryClass={cs.CV}
  }

@inproceedings{rajasegaran2023benefits,
    title={On the Benefits of 3D Pose and Tracking for Human Action Recognition},
    author={Rajasegaran, Jathushan and Pavlakos, Georgios and Kanazawa, Angjoo and Feichtenhofer, Christoph and Malik, Jitendra},
    booktitle={Proceedings of the IEEE/CVF Conference on Computer Vision and Pattern Recognition},
    pages={640--649},
    year={2023}
  }

@article{luvizon2020multi,
  title={Multi-task deep learning for real-time 3D human pose estimation and action recognition},
  author={Luvizon, Diogo C and Picard, David and Tabia, Hedi},
  journal={IEEE transactions on pattern analysis and machine intelligence},
  volume={43},
  number={8},
  pages={2752--2764},
  year={2020},
  publisher={IEEE}
}

@inproceedings{li2020detailed,
title={Detailed 2D-3D Joint Representation for Human-Object Interaction},
author={Li, Yong-Lu and Liu, Xinpeng and Lu, Han and Wang, Shiyi and Liu, Junqi and Li, Jiefeng and Lu, Cewu},
booktitle={CVPR},
year={2020}
}

@article{song7extrinsic,
  title={Extrinsic Calibration for Multiple Azure Kinect Cameras},
  author={Song, T},
  journal={GitHub repository. https://github. com/stytim/k4a-calibration (accessed Oct. 7, 2022.)}
}

@article{lsam,
  title={Language Segment-Anything},
  author={Luca, Medeiros},
  journal={GitHub repository. https://github.com/luca-medeiros/lang-segment-anything (accessed Feb. 15, 2025.)}
}

@article{riegel2016freecad,
  title={FreeCAD},
  author={Riegel, Juergen and Mayer, Werner and van Havre, Yorik},
  journal={Freecadspec2002. pdf},
  year={2016}
}

@inproceedings{multi-hmr2024,
    title={{Multi-HMR}: Multi-Person Whole-Body Human Mesh Recovery in a Single Shot},
    author={Baradel*, Fabien and 
            Armando, Matthieu and 
            Galaaoui, Salma and 
            Br{\'e}gier, Romain and 
            Weinzaepfel, Philippe and 
            Rogez, Gr{\'e}gory and
            Lucas*, Thomas
            },
    booktitle={ECCV},
    year={2024}
}

@inproceedings{liu2025core4d,
  title={CORE4D: A 4D Human-Object-Human Interaction Dataset for Collaborative Object REarrangement},
  author={Liu, Yun and Zhang, Chengwen and Xing, Ruofan and Tang, Bingda and Yang, Bowen and Yi, Li},
  booktitle={Proceedings of the Computer Vision and Pattern Recognition Conference},
  pages={1769--1782},
  year={2025}
}

@article{dosovitskiy2020vit,
  title={An Image is Worth 16x16 Words: Transformers for Image Recognition at Scale},
  author={Dosovitskiy, Alexey and Beyer, Lucas and Kolesnikov, Alexander and Weissenborn, Dirk and Zhai, Xiaohua and Unterthiner, Thomas and  Dehghani, Mostafa and Minderer, Matthias and Heigold, Georg and Gelly, Sylvain and Uszkoreit, Jakob and Houlsby, Neil},
  journal={ICLR},
  year={2021}
}

@inproceedings{TV_Human_Interaction,
   title = {High Five: Recognising human interactions in TV shows },
   author = {Patron, Alonso and Marszalek, Marcin and Zisserman, Andrew and Reid, Ian},
   year = {2010},
   pages = {50.1--50.11},
   booktitle = {BMVC},
   editors = {Labrosse, Fr\'ed\'eric and Zwiggelaar, Reyer and Liu, Yonghuai and Tiddeman, Bernie},
   isbn = {1-901725-40-5},
   note = {doi:10.5244/C.24.50}
}

@InProceedings{Hollywood2,
    author = "Marcin Marszalek and Ivan Laptev and Cordelia Schmid",
    title = "Actions in Context",
    booktitle = cvpr,
    year = "2009"
}

@inproceedings{shakefive,
 title = "Dyadic Interaction Detection from Pose and Flow",
 author = "Coert van Gemeren and Robby T. Tan and Ronald Poppe and Remco C. Veltkamp",
 booktitle={European Conference on Computer Vision (ECCV)},
    year={2014}}

@inproceedings{SBU_Kinect,
title={Two-person Interaction Detection Using Body-Pose Features and Multiple Instance Learning},
author={Kiwon Yun and Jean Honorio and Debaleena Chattopadhyay and Tamara L. Berg and Dimitris Samaras},
booktitle={Computer Vision and Pattern Recognition Workshops (CVPRW), 2012 IEEE Computer Society Conference on}, year={2012},
organization={IEEE}
}

@article{CMU_Panoptic,
  title={Panoptic Studio: A Massively Multiview System for Social Interaction Capture},
  author={Joo, Hanbyul and Simon, Tomas and Li, Xulong and Liu, Hao and Tan, Lei and Gui, Lin and Banerjee, Sean and Godisart, Timothy Scott and Nabbe, Bart and Matthews, Iain and Kanade, Takeo and Nobuhara, Shohei and Sheikh, Yaser},
  journal={IEEE Transactions on Pattern Analysis and Machine Intelligence},
  year={2017}
}

@INPROCEEDINGS{SALSA,  
author={Ricci, Elisa and Varadarajan, Jagannadan and Subramanian, Ramanathan and Bulò, Samuel Rota and Ahuja, Narendra and Lanz, Oswald},  booktitle={2015 ICCV},   title={Uncovering Interactions and Interactors: Joint Estimation of Head, Body Orientation and F-Formations from Surveillance Videos},   year={2015},  volume={},  number={},  pages={4660-4668},  doi={10.1109/ICCV.2015.529}}

@misc{Kinetics700,
      title={A Short Note on the Kinetics-700-2020 Human Action Dataset}, 
      author={Lucas Smaira and João Carreira and Eric Noland and Ellen Clancy and Amy Wu and Andrew Zisserman},
      year={2020},
      howpublished={arXiv:2010.10864},
}

@article{Moments,
  title={Moments in Time Dataset: one million videos for event understanding},
  author={Monfort, Mathew and Andonian, Alex and Zhou, Bolei and Ramakrishnan, Kandan and Bargal, Sarah Adel and Yan, Tom and Brown, Lisa and Fan, Quanfu and Gutfruend, Dan and Vondrick, Carl and others},
  journal={IEEE Transactions on Pattern Analysis and Machine Intelligence},
  year={2019},
  issn={0162-8828},
  pages={1--8},
  numpages={8},
  doi={10.1109/TPAMI.2019.2901464},
}

@misc{HACS,
title={HACS: Human Action Clips and Segments Dataset for Recognition and Temporal Localization},
author={Zhao, Hang and Yan, Zhicheng and Torresani, Lorenzo and Torralba, Antonio},
howpublished={arXiv:1712.09374},
year={2019}
          }

@misc{UT-Interaction,
      author = "Ryoo, M. S. and Aggarwal, J. K.",
      title = "{UT}-{I}nteraction {D}ataset, {ICPR} contest on {S}emantic {D}escription of {H}uman {A}ctivities ({SDHA})",
      year = "2010"
}

@INPROCEEDINGS{AVA,  author={Gu, Chunhui and Sun, Chen and Ross, David A. and Vondrick, Carl and Pantofaru, Caroline and Li, Yeqing and Vijayanarasimhan, Sudheendra and Toderici, George and Ricco, Susanna and Sukthankar, Rahul and Schmid, Cordelia and Malik, Jitendra},  booktitle=CVPR,   title={AVA: A Video Dataset of Spatio-Temporally Localized Atomic Visual Actions},   year={2018},  volume={},  number={},  pages={6047-6056},  doi={10.1109/CVPR.2018.00633}}

@InProceedings{rotation,
  title={Get the mean of the rotations},
  author={scipy, coders},
  journal={GitHub repository. https://github.com/scipy/scipy/blob/main/scipy/spatial/transform/_rotation.pyx (accessed Feb. 17, 2025.)}
}

@inproceedings{xie2023vistracker,
            title = {Visibility Aware Human-Object Interaction Tracking from Single RGB Camera},
            author = {Xie, Xianghui and Bhatnagar, Bharat Lal and Pons-Moll, Gerard},
            booktitle={IEEE Conference on Computer Vision and Pattern Recognition (CVPR)}, 
            month={June}, 
            year={2023} 
}

@inproceedings{
micikevicius2018mixed,
title={Mixed Precision Training},
author={Paulius Micikevicius and Sharan Narang and Jonah Alben and Gregory Diamos and Erich Elsen and David Garcia and Boris Ginsburg and Michael Houston and Oleksii Kuchaiev and Ganesh Venkatesh and Hao Wu},
booktitle={International Conference on Learning Representations},
year={2018},
url={https://openreview.net/forum?id=r1gs9JgRZ},
}

@INPROCEEDINGS{TIN-net,  author={Y. {Li} and S. {Zhou} and X. {Huang} and L. {Xu} and Z. {Ma} and H. {Fang} and Y. {Wang} and C. {Lu}},  booktitle=CVPR,  title={Transferable Interactiveness Knowledge for Human-Object Interaction Detection},   year={2019},  volume={},  number={},  pages={3580-3589},}

@INPROCEEDINGS{PMF,  author={B. {Wan} and D. {Zhou} and Y. {Liu} and R. {Li} and X. {He}},  booktitle=ICCV,  title={Pose-Aware Multi-Level Feature Network for Human Object Interaction Detection},   year={2019},  volume={},  number={},  pages={9468-9477},}
